%% file: bare_jrnl_new_sample4.tex
\documentclass[lettersize,journal]{IEEEtran}
\usepackage{amsmath,amsfonts}
\usepackage{algorithmic}
\usepackage{algorithm}
\usepackage{array}
\usepackage[caption=false,font=normalsize,labelfont=sf,textfont=sf]{subfig}
\usepackage{textcomp}
\usepackage{stfloats}
\usepackage{url}
\usepackage{verbatim}
\usepackage{graphicx}
\usepackage{cite}
\usepackage{tikz}
\usepackage{forest}
\usetikzlibrary{positioning,calc}
\usepackage[utf8]{inputenc}
\usepackage{booktabs}
\usepackage{multirow}
\usepackage{amssymb}
\usepackage{xcolor}

\DeclareRobustCommand{\benchtestonly}[1]{#1}
\DeclareRobustCommand{\benchwithtrain}[1]{#1}

\begin{document}

\title{Towards High-Level Semantic Intelligence}

\IEEEoverridecommandlockouts

\author{
Xiujie Song\textsuperscript{1},
Gefei Yang\textsuperscript{1},
Yining You\textsuperscript{1},
Jiahui Gan\textsuperscript{2},
Qi Jia\textsuperscript{3},\\
Shota Watanabe\textsuperscript{1},
Tianxi Wan\textsuperscript{1},
Mengyue Wu\textsuperscript{1,*},
and Kai Yu\textsuperscript{1}\\[3pt]
\textsuperscript{1}X-LANCE Lab, School of Computer Science, Shanghai Jiao Tong University, Shanghai, China\\
\textsuperscript{2}Nanjing University, Nanjing, China\\
\textsuperscript{3}Shanghai Artificial Intelligence Laboratory, Shanghai, China\\
\texttt{\{xiujiesong,mengyuewu\}@sjtu.edu.cn}
}

\maketitle

\begingroup
\renewcommand{\thefootnote}{\fnsymbol{footnote}}
\footnotetext[1]{Corresponding author: Mengyue Wu.}
\endgroup

\input{01_abstract.tex}
\input{02_intro.tex}

\input{03_theoretical_framework.tex}

\input{04_0_bls2hls.tex}

\input{04_1_high-level_semantic_intelligence_und.tex}

\input{04_2_high-level_semantic_intelligence_gen.tex}

\input{05_applications.tex}

\input{06_future.tex}

\input{07_conclusion.tex}

\bibliographystyle{IEEEtran}
\bibliography{reference}

\newpage

\vfill

\end{document}

%% file: 01_abstract.tex
\begin{abstract}

Recent advances in AI have substantially expanded its cognitive and reasoning capabilities. 
From the perspective of semantic complexity, the development of AI reveals a clear trajectory from simple to complex semantic processing. 
While early AI systems mainly addressed tasks involving direct and literal semantic perception or expression, contemporary systems are increasingly expected to perform more sophisticated cognitive reasoning, enabling the understanding and generation of High-Level Semantics (HLS). 
A similar trajectory can also be observed in human cognitive development. 
We define this transition as the shift from \emph{Basic-Level Semantic Intelligence (BLSI)} to \emph{High-Level Semantic Intelligence (HLSI)}. 
However, this issue has not yet been systematically and comprehensively examined in prior work. 
Motivated by this gap, this survey reviews the development of AI semantic intelligence from the perspective of semantic complexity. 
We systematically survey existing research on HLS tasks, including humor, sarcasm, metaphor, empathy, persuasion, narrative, and other general HLS phenomena, across text, speech, vision, and multimodal scenarios. 
Specifically, we summarize data construction methods, modeling and optimization strategies, and evaluation methodologies for both understanding and generation. 
HLS is essential for advancing AI toward genuinely human-like intelligence. 
By synthesizing existing methods and insights from the perspective of semantic intelligence, this survey aims to support the continued development of AI toward HLSI~\footnote{Working in progress. We maintain a real-time GitHub repository tracking progress at: \url{https://github.com/xiujiesong/Awesome-High-Level-Semantics}}.

\end{abstract}


%% file: 02_intro.tex
\section{Introduction}

\begin{figure}[htbp]
    \centering
    \includegraphics[width=0.49\textwidth]{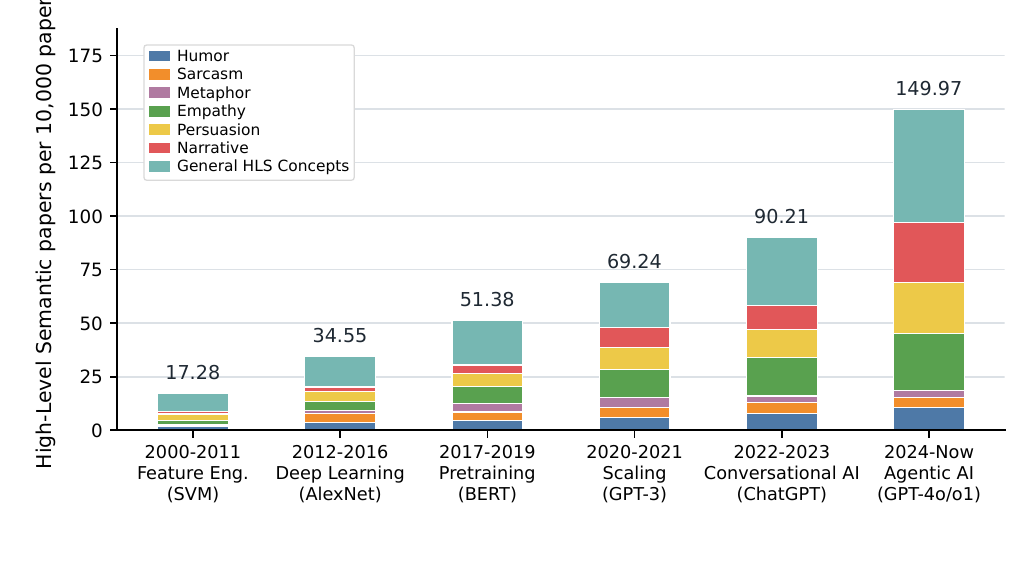}
    \caption{Evolutionary trajectory and publication density of High-Level Semantics (HLS) research across major AI venues (2000--Present). Statistics are compiled via the DBLP Search API across top-tier venues\protect\footnotemark[\value{footnote}]. The steady transition from feature engineering to modern agentic and reasoning paradigms underscores the shifting research emphasis toward complex semantic processing.} 
    \label{fig:paper_trend}
\end{figure}

In recent years, AI has undergone rapid progress, accompanied by a continuous expansion of its cognitive capabilities. 
In particular, its ability to process semantic information is undergoing a transition from recognition to cognition. 
Early AI systems primarily focused on tasks involving \textit{\textit{direct, literal}} perception or expression, requiring little complex reasoning, such as \textit{\textit{object recognition}}~\cite{726791}, \textit{\textit{speech pattern detection}}~\cite{1169179}, and \textit{\textit{language modeling}}~\cite{bengio2003neural}. 
With recent advances in AI systems, we now expect them to perform more complex forms of cognitive reasoning, including reasoning about visual causal relations, expressing emotions through vivid prosody, interpreting and generating metaphorical, humorous, and sarcastic content, responding empathetically, communicating persuasively, and constructing coherent narratives. 
As shown in Figure~\ref{fig:paper_trend}, these expanding expectations have been accompanied by growing research attention to complex semantic phenomena across successive stages of AI development.
In the 1940s, Shannon introduced information theory to quantify information in communication systems~\cite{6773024}, and Weaver interpreted communication in a broad sense as the process by which one mind may affect another~\cite{weaver2017recent}. 
In this new stage of AI progress, when we situate AI within the context of communication systems, this perspective inspires us to reconsider its evolution from a broader semantic standpoint. 
Viewed through this lens, AI is becoming increasingly capable of understanding and generating High-Level Semantics (HLS) through sophisticated cognitive reasoning. 
This development marks its ongoing transition from Basic-Level Semantic Intelligence (BLSI) toward High-Level Semantic Intelligence (HLSI).

\refstepcounter{footnote}

\footnotetext[\value{footnote}]{Full list includes NeurIPS, ICML, ICLR, ACL, EMNLP, NAACL, CVPR, ICCV, ECCV, ACM MM, AAAI, IJCAI, CHI, Interspeech, and ICASSP.}

\begin{figure*}[ht!] 
    \centering 
    \includegraphics[width=0.95\textwidth]{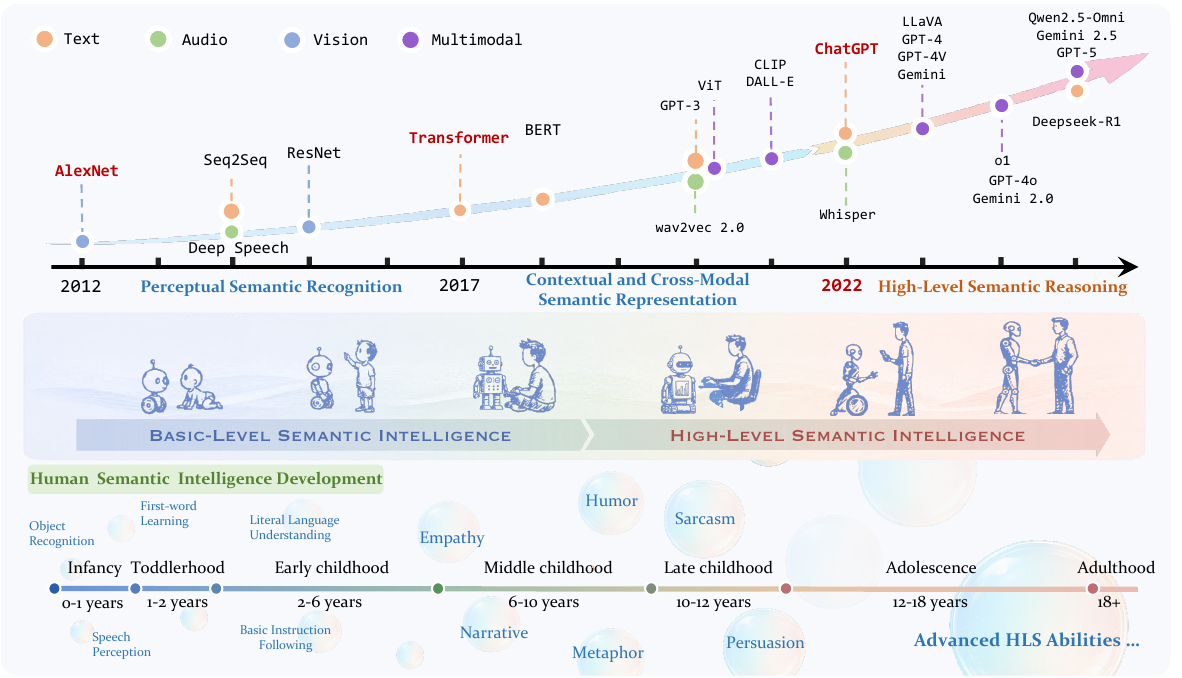} 
    \caption{Conceptual taxonomy and evolutionary roadmap from Basic-Level Semantic Intelligence (BLSI) to High-Level Semantic Intelligence (HLSI). Top: Milestones in machine semantic capabilities, progressing from early perceptual recognition (e.g., AlexNet, ResNet) to contextual representations (Transformer, BERT, CLIP) and high-level semantic reasoning (ChatGPT, GPT-4, DeepSeek-R1). Bottom: Parallel trajectories in human cognitive development, showing the progression from literal perception in infancy to expert-level abstract reasoning, humor, empathy, and narrative generation in adulthood.} 
    \label{fig:semantic_trend} 
\end{figure*}

Looking back at human development, we find that AI follows a strikingly similar trajectory in which semantic intelligence evolves from the processing of basic semantics toward increasingly complex semantics, as illustrated in Figure~\ref{fig:semantic_trend}. 
This trajectory is supported by evidence from cognitive science and developmental psycholinguistics, where early language acquisition is centered on lexical and conceptual meanings, whereas later stages involve more abstract, pragmatic, and socially embedded semantic phenomena~\cite{PINE_2005,bloom2002children}. 
Infants and toddlers first acquire word meanings and lexical–semantic relations grounded in familiar objects, actions, categories, and perceptual experience~\cite{PINE_2005,bloom2002children}. 
Later in development, children gradually acquire the ability to understand and produce higher-level semantics that go beyond surface meanings, including metaphor, irony, sarcasm, humor, empathy, narrative structure, and persuasion~\cite{winner1988point,airenti2016playing,hoffman1996empathy,bruner1990acts,stein1993theory}.
At the same time, these HLS abilities are not evenly mastered across individuals. They often depend on both innate talent and sustained learning, with advanced forms concentrated among a relatively small group of domain experts, such as artists, writers, psychological counselors, and actors. These expert-level human abilities represent the target capabilities of HLSI.

From the perspective of semantic intelligence, AI development can be broadly divided into three stages: perceptual semantic recognition, contextual and cross-modal semantic representation, and HLS reasoning, with the transitions marked by two milestone works: the Transformer~\cite{vaswani2017attention} and ChatGPT~\cite{achiam2023gpt}.
In the first stage, AI mainly focused on recognizing basic perceptual semantics, such as objects, attributes, actions, and categories. 
Representative works include early pattern recognition and expert systems~\cite{1096485}, as well as later statistical learning and deep learning models for image classification~\cite{NIPS2012_c399862d}, speech recognition~\cite{hannun2014deep}, and object detection~\cite{Girshick_2014_CVPR}, such as CNN-based visual recognition systems. 
The Transformer~\cite{vaswani2017attention} then enabled a shift toward contextual and cross-modal semantic representation, allowing models to capture meanings in relation to surrounding context and multimodal signals. 
This stage is represented by contextual language models such as BERT~\cite{47751} and GPT~\cite{radford2018improving}, vision-language models such as CLIP~\cite{pmlr-v139-radford21a}, and multimodal pretraining frameworks that align textual, visual, and other sensory representations~\cite{Sun_2019_ICCV,NEURIPS2021_cb3213ad}. 
Research at this stage focused on the basic understanding, generation, transformation, and alignment of textual, visual, and speech information, including tasks such as language modeling, machine translation, information extraction, visual question answering, speech recognition, text-to-image generation, and image–text alignment.
More recently, the rise of Large Language Models (LLMs) represented by ChatGPT, along with the subsequent development of Multi-modal Large Language Models (MLLMs), has brought breakthrough advances in AI’s basic language and multimodal capabilities~\cite{alayrac2022flamingo,NEURIPS2023_6dcf277e,achiam2023gpt}. 
At this stage, we argue that AI has largely mastered basic semantic abilities and is progressively moving toward HLS reasoning~\cite{bubeck2023paper}. 
Accordingly, we regard ChatGPT as a rough boundary between BLSI and HLSI. 
This shift is also reflected in Figure~\ref{fig:paper_trend} by the pronounced growth of HLS research during the conversational and agentic AI eras. 
Though they have already acquired basic capabilities in recognition, reasoning, and generation, they still face significant challenges in tasks that require the understanding and generation of complex semantics. 
A clear example can be seen in the recently popular field of video generation. 
Although AI is now capable of generating realistic videos to tell a story, the semantic-level design of the story and other intended semantic effects still largely depends on human input~\cite{opensora2024, seedance2026seedance}. 
This reveals the current limitations of AI in semantic-level generation. 
Therefore, a systematic review of the current development of HLS tasks is needed to clarify the key challenges, methodological directions, and future opportunities for advancing AI toward HLSI.

Motivated by this observation, this survey revisits recent progress in AI through the lens of semantic intelligence. 
We argue that HLS should be treated not merely as a collection of isolated downstream tasks, but as a coherent research direction concerned with how AI systems understand, generate, and reason about complex semantics. 
To this end, we review representative forms of HLS, summarize existing computational methods, and identify key challenges in modeling their underlying cognitive and reasoning complexity. 
Specifically, in this survey, we discuss the problems and methods of HLS research through six well-defined HLS categories: \textit{humor, sarcasm, metaphor, empathy, persuasion, and narrative}, as well as other general HLS domains\footnote{
Note that we exclude abilities such as mathematical reasoning and code generation from our scope, as their meanings are primarily governed by explicit formal rules, whereas the HLS studied here are context-dependent, socially grounded, and shaped by human intentions.}. 
Among related works, Song et al.~\cite{song2025large} reviews subjective-language understanding tasks such as sentiment, emotion, figurative language, stance, intent, and aesthetics, while Mago et al.~\cite{BeyondtheObvious2026} discusses abstract concepts and high-level semantics in video understanding. 
However, neither provides a comprehensive account of HLSI from the fundamental perspective of semantic complexity, across modalities such as text, speech, and vision, and across both understanding and generation tasks. 
Ultimately, we believe that HLSI constitutes a crucial step toward aligning AI with human cognitive capabilities, as it requires machines to move beyond surface-level pattern recognition toward contextual understanding, intention-sensitive reasoning, and meaningful human-centered communication.

The contributions of this survey are summarized as follows:
\begin{itemize}
    \item To the best of our knowledge, this is the first comprehensive survey of HLSI through the lens of semantic complexity, covering both understanding and generation across text, speech, vision, and multimodal settings.
    
    \item We establish a conceptual framework that distinguishes BLSI from HLSI, characterizes the core properties of HLS, and explains how HLS is formed, understood, and generated (Section~\ref{hls:theory}).
    
    \item We organize the key problems and tasks associated with BLSI and HLSI, characterize HLS tasks from the perspectives of their inputs and outputs, and illustrate the broader transition from BLSI toward HLSI (Section~\ref{sec:bls2hls}).
    
    \item For both HLS understanding and generation, we systematically review existing data construction methods, modeling and optimization approaches, and evaluation methodologies (Sections~\ref{sec:hlsu} and~\ref{sec:hlsg}).
    
    \item We discuss the applications enabled by HLSI and identify key challenges and promising directions for future research (Sections~\ref{sec:applications} and~\ref{sec:future_directions}). 
\end{itemize}

%% file: 03_theoretical_framework.tex
\section{Theoretical Framework}
\label{hls:theory}

To systematically analyze HLSI, we establish a formal theoretical framework grounded in semantic complexity and cognitive reasoning. 
In this section, we formalize the foundational elements of HLS, define formal metrics for semantic complexity and density, and delineate the structural progression from BLSI to HLSI.

\subsection{Formal Definitions}
\label{sec:def}

\subsubsection{Basic Semantic Elements}

Although HLS is inherently abstract, we seek to provide a structured account of how it can be formally represented and computationally modeled.
Specifically, we characterize HLS through four fundamental elements: Semantic Clues, Semantic Chains, Causal Chains, and Semantic Effects.
These elements respectively capture the evidential basis, inferential process, causal structure, and holistic cognitive, affective, or communicative effect underlying HLS interpretation. 

\paragraph{Semantic Clue}

Semantic Clues ($c$) serve as the foundational elements of semantics and constitute the prerequisite for subsequent cognitive reasoning in HLS understanding. 
The acquisition of semantic clues largely depends on perception ability, since a system must first identify relevant clues before reasoning over them.
Therefore, accurate semantic clue recognition is essential, as missing or noisy clues may mislead reasoning and cause semantic deviation.

\paragraph{Semantic Chain}

A Semantic Chain (SC) serves to associate semantic clues, enabling the derivation of higher semantic conclusions via cognitive reasoning. 
In this paper, we refer to this phenomenon as \textit{Semantic Level-Up}, as shown in Figure~\ref{fig:semantic_up}. 
It can be expressed as
\begin{equation}
\{c_1,c_2,\ldots,c_n\}\vdash y,
\end{equation}
where $c_1, c_2, \ldots, c_n$ denote semantic clues, $y$ denotes the derived HLS conclusion, and $\vdash$ denotes a semantic inference relation, indicating that $y$ can be derived by jointly interpreting the clues in light of relevant contextual and background knowledge.

\paragraph{Causal Chain}

HLS often encode rich causal relationships, which we represent using the Causal Chain (CC).
A Causal Chain serves to represent causal relations within semantics. 
It can be expressed as
\begin{equation}
\{a_1, a_2, \cdots, a_n\} \rightsquigarrow e,
\end{equation}
where $a_1, a_2, \ldots, a_n$ denote a set of causal factors, $e$ denotes the resulting effect, and $\rightsquigarrow$ denotes a causal relation, indicating that the joint influence of these factors gives rise to or contributes to $e$.

\paragraph{Semantic Effect}

A Semantic Effect (SE) refers to the holistic cognitive, affective, or communicative effect that emerges from the joint composition of Semantic Chains and Causal Chains.
While Semantic Chains associate semantic clues to derive high-level meanings, Causal Chains capture how relevant factors, events, and mental states give rise to particular outcomes.
Their interaction forms a coherent semantic structure that can produce effects such as humor, emotional resonance, sarcasm, empathy, and persuasion.
Formally, a Semantic Effect can be represented as
\begin{equation}
\mathcal{E}
=
\Phi\left(
\{\mathrm{SC}_i\}_{i=1}^{p},
\{\mathrm{CC}_j\}_{j=1}^{q}
\right),
\end{equation}
where $\mathcal{E}$ denotes the resulting Semantic Effect, $\mathrm{SC}_i$ and $\mathrm{CC}_j$ denote the involved Semantic Chains and Causal Chains, respectively, and $\Phi$ denotes their semantic composition.
Therefore, a Semantic Effect is not determined by an isolated clue or relation, but emerges from the coordinated interpretation of multiple semantic and causal structures.

\subsubsection{Semantic Measurement}

HLS can be quantitatively characterized from two complementary perspectives: Semantic Complexity, which reflects the reasoning effort required to derive meaning, and Semantic Density, which captures how much semantic complexity is encoded within a given physical representation. 

\paragraph{Semantic Complexity ($C_s$)}

Semantic Complexity measures the intricacy of the reasoning process required to derive meaning $S$ from a set of semantic clues $\mathcal{C}$. 
High semantic complexity implies that multiple layers of inference, causal reasoning, and contextual integration are required. 
It is therefore often associated with longer multi-hop reasoning paths involving a greater number of inferential steps. 
In a simplified form, semantic complexity can be roughly characterized by the complexity of the Semantic Chains and Causal Chains involved:
\begin{equation}
C_s \propto \mathrm{Complexity}(\mathrm{SC}, \mathrm{CC}).
\end{equation}

\paragraph{Semantic Density ($D_s$)}
Semantic Density is defined as the ratio of Semantic Complexity ($C_s$) to the physical representation size ($R$) (e.g., token count or pixel area). 
HLS often exhibits high density, conveying profound meaning in minimal representation.

\begin{equation}
    D_s = \frac{C_s}{|R|}
\end{equation}

\subsubsection{Basic-Level Semantic Intelligence and High-Level Semantic Intelligence}

\paragraph{Basic-Level Semantic Intelligence (BLSI)}

\textbf{Basic-Level Semantics (BLS)} is typically literal, with a relatively direct mapping between perceptual observation and meaning. 
\textbf{BLSI} denotes the capacity to process such perception-grounded semantic information that supports routine communication. 
It serves as the foundation and prerequisite for HLSI. 
BLS understanding refers to the ability to comprehend literal meanings, standard instructions, and factual descriptions from multimodal inputs, where their semantic interpretation is directly grounded in perceptual or contextual clues. 
BLS generation refers to the ability to generate coherent, well-formed, and internally consistent outputs across modalities that convey conventional semantic content. 
It further includes the transformation of BLS across modalities, where literal semantic content is preserved while its representational form changes.

\begin{figure*}[htbp] % h:当前位置, t:页顶, b:页底, p:独立一页
    \centering % 图片居中
    \includegraphics[width=0.9\textwidth]{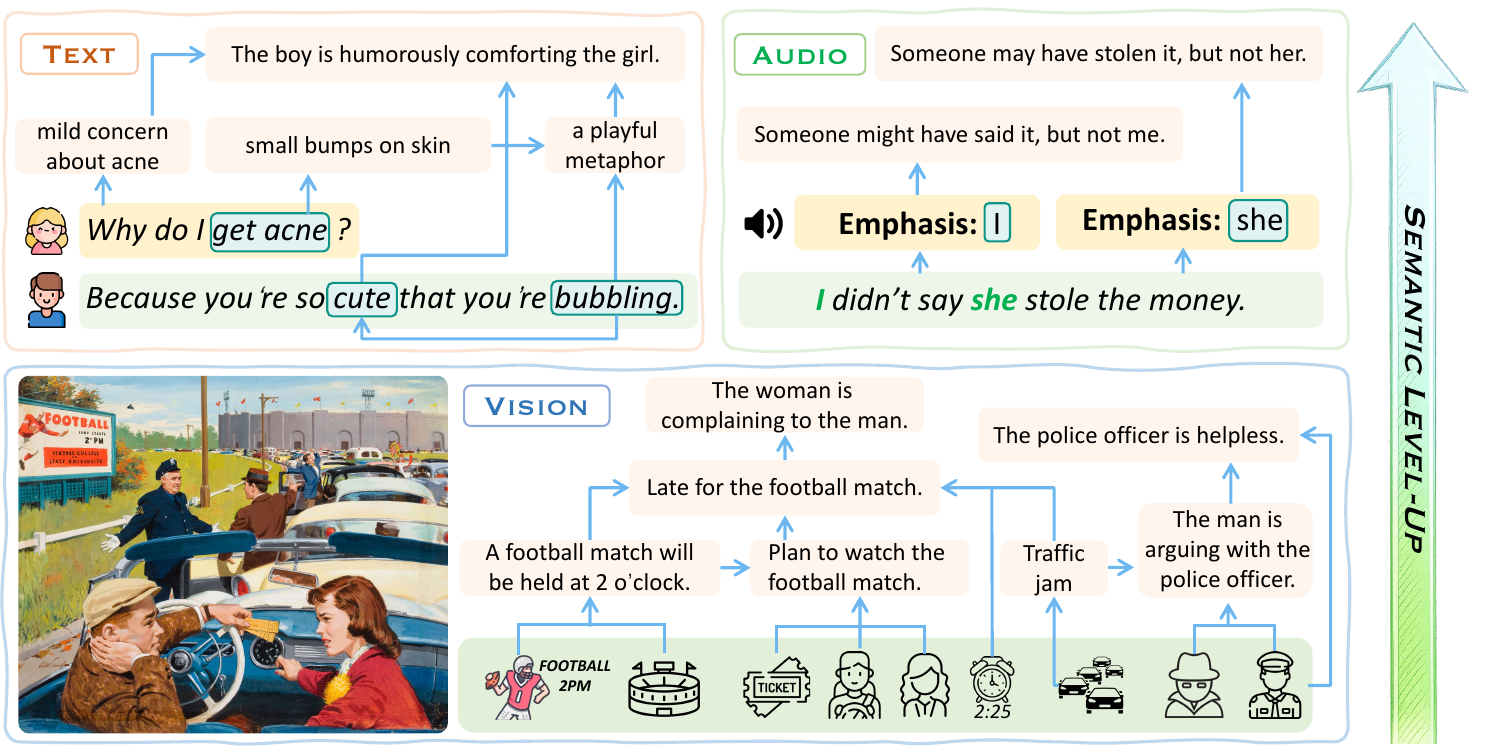} % 插入图片，设置宽度为正文宽度的80%
    \caption{Illustration of Semantic Level-Up through cognitive reasoning across different modalities. By combining lower-level perceptual or contextual semantic clues, systems perform cognitive reasoning to derive higher-level semantic conclusions. 1) For Text, combining literal statements about skin conditions leads to a playful, humorous metaphor. 2) For Audio, shifting vocal stress across the same sentence alters the implied non-literal meaning and emphasis. 3) For Vision, integrating observations such as arguments, police presence, traffic conditions, schedule tickets, clock times, and emotional expressions supports the construction of higher-level causal narratives and the interpretation of complex event dynamics and social interactions.
    } 
    \label{fig:semantic_up} % 用于正文引用
\end{figure*}

\paragraph{High-Level Semantic Intelligence (HLSI)}

\textbf{High-Level Semantics (HLS)} refers to abstract meaning constructed through the integration of basic-level observations and complex cognitive reasoning.
For example, as illustrated in Figure~\ref{fig:semantic_up}, we need not only to recognize basic textual, auditory, and visual information, but also to connect these information units through cognitive reasoning in order to ultimately understand these HLS phenomena. 
HLS encompasses a broad range of complex semantic phenomena, including context-dependent, intent-sensitive, and socially situated meanings commonly studied in computational pragmatics, while also extending to broader, holistic semantic effects that emerge across an entire discourse, interaction, or multimodal experience, such as cognitive, affective, and narrative effects. 
\textbf{HLSI} refers to the capacity to comprehend and generate HLS through higher-order cognitive processes. 
Specifically, HLS understanding requires the integration of perception, knowledge, and reasoning to infer latent meanings. HLS generation likewise relies on these capabilities, while also drawing on creative and innovative processes to construct content that conveys intended high-level meanings. 
HLSI aims to model the level of semantic intelligence exhibited by domain experts such as writers, artists, psychological counselors, and master strategists. 
In this survey, we focus on six well-defined and systematically studied HLS tasks—metaphor, humor, sarcasm, empathy, persuasion, and narrative—while also covering several general, cross-cutting tasks that span these specific phenomena.

\subsection{Characteristics of High-Level Semantics}

Drawing on the preceding definition and a synthesis of existing studies on the HLS tasks covered in this survey, we summarize several recurring characteristics of HLS. 

\paragraph{Ubiquity Across Modalities}

Semantic information is inherently conveyed through multimodal media, making HLS ubiquitous across modalities capable of carrying semantic information. 
Although HLS is multimodal in nature, each individual modality also exhibits a progression toward more complex HLS as illustrated in Figure~\ref{fig:semantic_up}.

\paragraph{Beyond Literal Meaning}

HLS goes beyond the recognition of observable signals in the input. 
It requires models to reason over these signals and infer meanings beyond the literal surface. 
This also contributes to the abstract nature of HLS.

\paragraph{Cognitive Complexity}

HLS typically involves a higher degree of semantic complexity, as its understanding and generation often requires multi-hop reasoning. 
Consequently, processing HLS demands substantially greater cognitive effort.

\paragraph{Emotional Arousal}

HLS often gives rise to semantic effects, which may further elicit emotional arousal in the receiver. 
Therefore, emotional responses also serve as one of the criteria for assessing the semantic effects of HLS.

\paragraph{Abstraction}
A key characteristic of HLS is abstraction. 
Unlike BLS, which is often grounded in directly observable entities, attributes, actions, or explicit linguistic meanings, HLS usually refers to meanings that are not fully specified by surface forms. Such meanings are often implicit, context-dependent, and cognitively constructed. 

\paragraph{Subjective}

Since HLS is constructed by receivers through cognitive reasoning, its interpretation is inevitably influenced by individual differences in knowledge, experience, cultural background, emotional state, and reasoning ability. 
As a result, the semantic effects of HLS often exhibit subjectivity across different receivers.

\subsection{Understanding and Generating High-Level Semantics}

\begin{figure}[htbp] % h:当前位置, t:页顶, b:页底, p:独立一页
    \centering % 图片居中
    \includegraphics[width=0.49\textwidth]{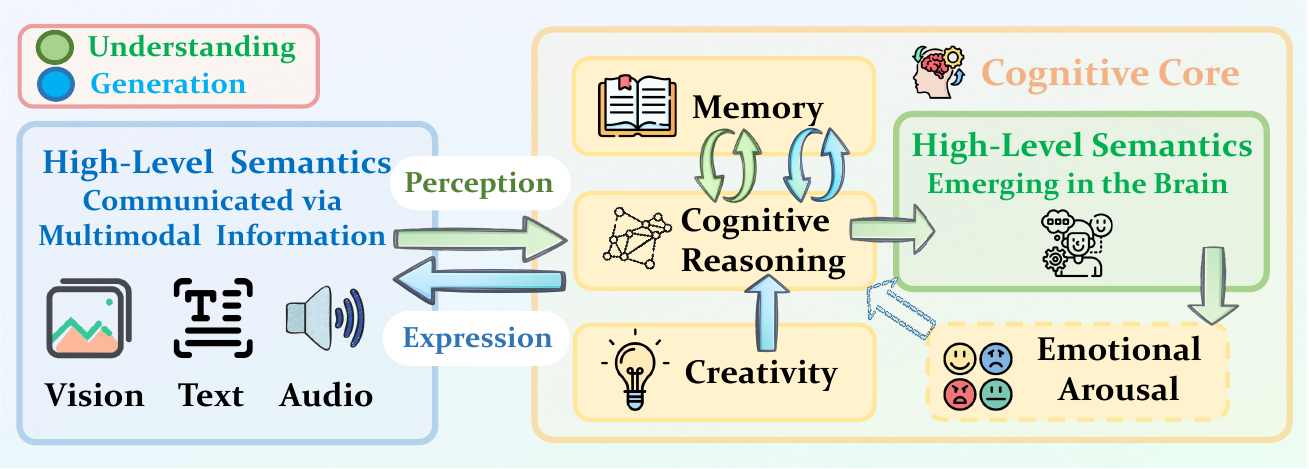} % 插入图片，设置宽度为正文宽度的80%
    \caption{The cognitive architecture of High-Level Semantic Understanding and Generation. During understanding, multimodal physical signals (text, vision, audio) undergo perceptual processing and brain-level cognitive reasoning—integrating background knowledge and context from memory—to form abstract semantic comprehension and elicit emotional arousal. During generation, cognitive reasoning, memory retrieval, human creativity, and emotional drive synthesize high-level semantic intent, which is then expressed outward into the physical world via multimodal channels.} % 图片下方的文字说明
    \label{fig:understand_generate_hls} % 用于正文引用
\end{figure}

Figure~\ref{fig:understand_generate_hls} illustrates the processes of HLS understanding and generation. 
In this paper, we distinguish understanding from generation based on whether a task identifies or explains existing semantics or intended effects, or produces new semantics or effects through generation.

HLS is often transmitted in the physical world in multimodal forms.
From the perspective of semantic understanding, humans first receive multimodal signals through fundamental perceptual abilities, including text, vision, and audio perception. 
These signals are then processed in the brain through complex cognitive reasoning. 
During this reasoning process, the brain often retrieves relevant background knowledge and contextual information from memory, which further contributes to the formation of semantic understanding. Ultimately, HLS emerges in the brain, and such semantic understanding may subsequently induce emotional responses in humans.

In the process of HLS generation, the brain likewise relies on complex cognitive reasoning and knowledge retrieval mechanisms. 
Unlike semantic understanding, semantic generation additionally requires creative capabilities and may sometimes be influenced by emotions. 
Through these processes, the intended semantics are formed and finally expressed to the external world through different modalities.

%% file: 04_0_bls2hls.tex
\section{From Basic-level Semantic Intelligence to High-Level Semantic Intelligence}
\label{sec:bls2hls}

In this section, we introduce the main research problems of BLSI and HLSI from the perspective of tasks. 

\subsection{Basic-Level Semantic Tasks}

BLS tasks provide essential semantic grounding for downstream high-level capabilities. 
In \textbf{natural language processing}, representations have evolved from surface-level co-occurrence statistics~\cite{brown-etal-1992-class} to distributed embeddings~\cite{mikolov2013efficient}, and further to Transformer-based architectures~\cite{vaswani2017attention} representing each token in relation to its surrounding context. 
Based on these representations, early NLP tasks focused on processing explicit textual information, including named entity recognition~\cite{tjong-kim-sang-de-meulder-2003-introduction}, information extraction~\cite{grishman-sundheim-1996-message}, machine translation~\cite{koehn2003statistical,DBLP:journals/corr/BahdanauCB14}, and text understanding tasks such as reading comprehension~\cite{rajpurkar-etal-2016-squad}, summarization~\cite{rush-etal-2015-neural}, and paraphrase identification~\cite{dolan-brockett-2005-automatically}. 
These tasks mainly concern explicit semantic content, rather than implicit beliefs, communicative intentions, or abstract ideas. 
In \textbf{computer vision}, representation has progressed from pixel-level pattern recognition to increasingly stable semantic representations through deep convolutional networks trained on large-scale datasets~\cite{726791, 5206848, NIPS2012_c399862d}. 
Correspondingly, vision tasks have expanded recognition from whole images to more detailed descriptions of visible content, such as object detection~\cite{NIPS2015_14bfa6bb}, semantic and instance segmentation~\cite{Long_2015_CVPR},~\cite{he2017mask}, and pose estimation~\cite{cao2017realtime}, alongside generative tasks including texture synthesis~\cite{wei2009state,gatys2015texture}, image completion~\cite{iizuka2017globally,sun2005image}, and data-driven image generation with variational autoencoders~\cite{kingma2013auto} and generative adversarial networks~\cite{goodfellow2014generative}. 
These tasks characterize what is present and how it is spatially organized, but remain grounded in observable entities and layouts. 
In \textbf{audio processing}, representations have advanced from statistical temporal modeling~\cite{18626} with hidden Markov models to speech enhancement and denoising methods such as spectral subtraction~\cite{1163209}, then further to end-to-end ASR systems with mappings from acoustic signals to characters or words~\cite{hannun2014deep,7472621}. 
Early audio tasks, including automatic speech recognition~\cite{18626}, speaker recognition~\cite{8461375}, speech synthesis and text-to-speech tasks~\cite{schroder2001emotional,tokuda2000speech}, as well as later neural waveform or spectrogram generators including WaveNet~\cite{van2016wavenet} and Tacotron~\cite{wang2017tacotron}, focus on converting continuous waveforms into linguistic,
speaker-related, or event-level descriptions, primarily addressing recognizable acoustic content. 
With the rapid development of AI, research is increasingly shifting toward cross-modal alignment~\cite{pmlr-v139-radford21a}, interaction~\cite{alayrac2022flamingo}, and unified representation learning ~\cite{girdhar2023imagebind}, where basic-level semantic tasks across text, visual, and audio modalities provide the necessary foundation for this transition by transforming raw signals into structured semantic representations.

\begin{table*}[th!]
\centering
\caption{
Comprehensive summary of representative datasets and benchmarks across six major High-Level Semantic (HLS) categories (Humor, Metaphor, Sarcasm, Empathy, Persuasion, and Narrative) and other general HLS tasks, categorized by modalities: Text, Vision, Audio, and Multimodality. 
} 
\label{table:datasets_simple}
\small
\renewcommand{\arraystretch}{1.5} % 稍微加高行距，方便阅读
\begin{tabular}{lp{12.5cm}} % 删去一列，增加数据集列的宽度
\toprule
\textbf{Category} & \multicolumn{1}{c}{\textbf{Datasets / Benchmarks}} \\
\midrule

\textbf{Humor} & 
\textbf{[Text]} MCL~\cite{mihalcea2005making}, \#HashtagWars~\cite{potash-etal-2017-semeval}, Humicroedit~\cite{hossain-etal-2019-president}, CAH~\cite{ofer-shahaf-2022-cards}, $C^3$~\cite{li-etal-2023-language}, TalkFunny~\cite{Chen_Yuan_Liu_Liu_Guan_Guo_Peng_Liu_Li_Xiao_2024}, \benchtestonly{Chumor 2.0}~\cite{he-etal-2025-chumor},  PHUNNY~\cite{cocchieri-etal-2025-call}, CFunSet~\cite{yu2025cfunmodel},  Comparing Apples to Oranges~\cite{loakman-etal-2025-comparing}, \benchtestonly{DrivelHub}~\cite{wang2025drivel} \newline
\textbf{[Vision]} \benchtestonly{New Yorker Caption Contest}~\cite{hessel-etal-2023-androids}, \benchwithtrain{HumorousAI}~\cite{zhang2024humor}, Oogiri-GO~\cite{a78fdd6694d54f5491f7393dfc7529e1}, \benchtestonly{YESBUT}~\cite{hu2024cracking}, \benchtestonly{VisionArena}~\cite{11094422}, \benchtestonly{D-HUMOR}~\cite{kasu2025d}, \benchtestonly{PixelHumor}~\cite{ryan-etal-2025-humor}, \benchtestonly{HumorDB}~\cite{jain2025humordb}, \benchtestonly{v-HUB}~\cite{shi2025v},  
\benchtestonly{GODBench}~\cite{lei-etal-2025-godbench}\newline
\textbf{[MultiModality]} ExFunTube~\cite{ko-etal-2023-language}, Passau-SFCH~\cite{10707160}, StandUp4AI~\cite{barriere-etal-2025-standup4ai}\\
\midrule

\textbf{Metaphor} & 
\textbf{[Text]} \benchtestonly{Fig-QA}~\cite{liu-etal-2022-testing}, \benchtestonly{FLUTE}~\cite{chakrabarty-etal-2022-flute}, CMRE~\cite{chen-etal-2023-chinese}, CIP~\cite{qiang2023chinese}, MABL~\cite{kabra-etal-2023-multi}, NewsMet~\cite{joseph-etal-2023-newsmet}, MMTE~\cite{wang-etal-2024-mmte}, VMC-P~\cite{mao-etal-2024-metapro}, \benchtestonly{MUNCH}~\cite{tong-etal-2024-metaphor}, METORIE~\cite{10890213}, FLUTE.st~\cite{sengupta2025investigating}, Metaphoric Analogies~\cite{boisson-etal-2025-automatic}, MABL\&MAPS~\cite{khoshtab-etal-2025-comparative} \newline
\textbf{[Vision]} HAIVMet~\cite{chakrabarty-etal-2023-spy}, \benchtestonly{MultiCMET}~\cite{zhang-etal-2023-multicmet}, MEMECAP~\cite{hwang-shwartz-2023-memecap}, MultiMM~\cite{yang-etal-2025-cultural}, \benchtestonly{InfoChartQA}~\cite{NEURIPS2025_c7bd72f3}, EmoMeta~\cite{lu2025emometa}, CM3D~\cite{zhang2025towards}, \benchtestonly{Visual Puzzles}~\cite{lee-etal-2025-puzzled}, BANMIME~\cite{mia-etal-2025-banmime},  ImageMet~\cite{kundu2025looking} \newline
\textbf{[Audio]} Unspoken~\cite{xiao2025can}    \\
\midrule

\textbf{Sarcasm} & 
\textbf{[Text]} \benchtestonly{iSarcasmEval}~\cite{abu-farha-etal-2022-semeval}, CSC~\cite{jang-frassinelli-2024-generalizable}, SarcasmBench~\cite{11146812}, MultiPICo~\cite{casola-etal-2024-multipico}, RedSD~\cite{hong-etal-2025-rhetorical}, \benchwithtrain{FanChuan}~\cite{zheng-etal-2025-fanchuan} \newline
\textbf{[Vision]} MuSG~\cite{zhao-etal-2023-multi}, SarcNet~\cite{yue-etal-2024-sarcnet},  \benchwithtrain{DocMSU}~\cite{Du_Nan_Zhang_Xie_Xu_Fan_Cui_Tao_Jiang_2024}, MIMOSA~\cite{ahsan-etal-2024-multimodal}, \benchtestonly{YesBut}~\cite{nandy-etal-2024-yesbut}, MCDSCS~\cite{10.1145/3664647.3680978}, BHM~\cite{hossain-etal-2024-deciphering},  \benchwithtrain{MMSD3.0}~\cite{zhao2026mmsd3} \newline
\textbf{[Audio]} \benchtestonly{Speech-Specific Risk}~\cite{yang-etal-2024-towards-probing}, ToxicTone~\cite{luo2025toxictonemandarinaudiodataset}, PolyHope V2~\cite{butt2025optimismexpectationsarcasmmulticlass},  PodSarc~\cite{li2026leveraginglargelanguagemodels}  \newline
\textbf{[MultiModality]} MUStARD~\cite{castro-etal-2019-towards}, WITS~\cite{kumar-etal-2022-become}, MCSD~\cite{gao25f_interspeech}
\\
\midrule

\textbf{Empathy} & 
\textbf{[Text]} MPED~\cite{zhu-etal-2022-multi}, EQT~\cite{svikhnushina-etal-2022-taxonomy}, PAL~\cite{mishra-etal-2023-pal}, \benchtestonly{EmotionBench}~\cite{huang2024apathetic}, HEART-felt Stories~\cite{shen-etal-2024-heart}, ALOE~\cite{yang-jurgens-2024-modeling}, \benchtestonly{EmotionQueen}~\cite{chen-etal-2024-emotionqueen}, SYNTHEMPATHY~\cite{chen2025synthempathy}, ECC~\cite{he2025ecc}, EmpathyFromPerspectives~\cite{chen-etal-2025-empathy}, TIDE~\cite{bn2025pursuit}, Empathy-QA~\cite{yao2025empathy}, EmoCare~\cite{shi-etal-2025-beyond}, \benchtestonly{TactfulToM}~\cite{liu-etal-2025-tactfultom},   SENSE-7~\cite{suh2026sense}, \benchwithtrain{KardiaBench}~\cite{10.1145/3774904.3793022} \newline
\textbf{[Vision]} STICKERCONV~\cite{zhang-etal-2024-stickerconv}, EmpathyAgent~\cite{chen2025empathyagent} \newline
\textbf{[Audio]} NCSSD~\cite{liu2024generative}, EChat-200K~\cite{geng2025osum}, 	\benchtestonly{EchoMind}~\cite{zhou2025echomindinterrelatedmultilevelbenchmark}, \benchtestonly{AEQ-Bench}~\cite{luo2026aeq} \newline % EmpatheticVideos~\cite{chen2024detecting}, 
\textbf{[MultiModality]} MEDIC~\cite{zhu2023medic}, EmpathicStories++~\cite{shen-etal-2024-empathicstories} \\
\midrule

\textbf{Persuasion} & 
\textbf{[Text]} DailyPersuasion~\cite{jin-etal-2024-persuading}, \benchtestonly{AmazonHistoryPrice}~\cite{xia-etal-2024-measuring}, \benchtestonly{NegotiationToM}~\cite{chan2024negotiationtom}, SafePersuasion~\cite{kong-etal-2025-safepersuasion}, \benchtestonly{PersuasiveToM}~\cite{yu2025persuasivetom}, DCN~\cite{wang2025debt}, \benchtestonly{Persuade Me If You Can}~\cite{za2025persuade}, \benchtestonly{CToM-Persu}~\cite{zhang2026persuasion} \newline
\textbf{[Vision]} Persuasion Strategy Corpus~\cite{aaai23advertisements},  \benchtestonly{SemEval-2024 Task 4}~\cite{dimitrov-etal-2024-semeval}, PVP~\cite{kim-etal-2025-pvp}\\
\midrule

\textbf{Narrative} & 
\textbf{[Text]}  ROCStories~\cite{mostafazadeh-etal-2016-corpus}, Event2Mind~\cite{rashkin-etal-2018-event2mind}, NarrativeQA~\cite{kocisky-etal-2018-narrativeqa}, Story Commonsense~\cite{rashkin-etal-2018-modeling}, LiSCU~\cite{brahman-etal-2021-characters-tell}, FairytaleQA~\cite{xu-etal-2022-fantastic}, \benchtestonly{LOT}~\cite{guan-etal-2022-lot}, NOCHA~\cite{karpinska-etal-2024-one}, DetectBench~\cite{gu-etal-2024-detectbench}, StorySeeker~\cite{antoniak-etal-2024-people}, DetectiveQA~\cite{xu2025detectiveqaevaluatinglongcontextreasoning}, \benchtestonly{PRELUDE}~\cite{yu2025preludebenchmarkdesignedrequire}, \benchtestonly{NOVELHOPQA}~\cite{gupta-etal-2025-novelhopqa}, 
CHATTER~\cite{baruah-narayanan-2025-chatter}, 
\benchtestonly{WHODUNIT}~\cite{gupta2025whodunitevaluationbenchmarkculprit}, FLAWEDFICTIONS~\cite{ahuja2025finding}, \benchtestonly{TurnaboutLLM}~\cite{yuan-etal-2025-turnaboutllm} \newline % , \benchtestonly{TLDM}~\cite{hamilton-etal-2026-long-didnt}
\textbf{[Vision]}  STRIPCIPHER~\cite{wang2025beyond}, PopCaptions~\cite{sachdeva2025panels}, \benchtestonly{CogBench}~\cite{song-etal-2025-cognitive}, ISA~\cite{Song_Pang_Tang_Wu_Zhu_2025}, \benchtestonly{\ensuremath{R^3}-VQA}~\cite{niu2025r}, \benchtestonly{V-SOCIAL}~\cite{lin-etal-2025-v}, \benchtestonly{MOMENTS}~\cite{villa-cueva-etal-2025-moments}, \benchtestonly{SeriesBench}~\cite{Zhang_2025_CVPR}, \benchtestonly{VRBench}~\cite{yu2025vrbench},   \benchtestonly{SAGAQA}~\cite{pennec2026sagaqamultihopreasoningbenchmark}, \benchtestonly{NARRATIVETRACK}~\cite{ha2026narrativetrackevaluatingentitycentricreasoning}, ComicVQA~\cite{gan-etal-2026-comicvqa} 
  \newline
\textbf{[MultiModality]} 
 Movie101v2~\cite{yue-etal-2025-movie101v2} \\
\midrule

\textbf{Others} & 
\textbf{[Text]} CUP~\cite{sun2022context}, ExPUNations~\cite{sun-etal-2022-expunations}, PIE-English~\cite{adewumi-etal-2022-potential}, ANALOGICAL~\cite{wijesiriwardene-etal-2023-analogical}, MMFL~\cite{lai-etal-2023-multilingual}, DiPlomat~\cite{li2023diplomat}, \benchwithtrain{SOCKET}~\cite{choi-etal-2023-llms}, \benchtestonly{Cskills}~\cite{zhou-etal-2024-think}, ChinesePun~\cite{chen-etal-2024-u}, SLANG~\cite{mei-etal-2024-slang},  \benchtestonly{FLUB}~\cite{li2024llms}, PiC~\cite{wang-etal-2025-proverbs}, 
 \benchtestonly{FLUID QA}~\cite{park2025fluid}, ToMEmoReason~\cite{yeo2025beyond},
AHaPairs~\cite{kim-etal-2026-kind},
\benchtestonly{RedTrans-Bench}~\cite{guo2025redefining},  \benchtestonly{Pun2Pun}~\cite{ma-etal-2025-pun2pun}, \benchtestonly{MENT}~\cite{tian2026literalmappingbenchmarkingimproving},  \benchtestonly{CHEER}~\cite{huang-etal-2025-large}  \newline
\textbf{[Vision]}  FigMemes~\cite{liu-etal-2022-figmemes}, SMILE~\cite{hyun-etal-2024-smile}, \benchtestonly{DEEPEVAL}~\cite{yang-etal-2024-large}, \benchtestonly{II-Bench}~\cite{liu2024ii}, \benchtestonly{V-FLUTE}~\cite{saakyan-etal-2025-understanding}, \benchtestonly{InsightVision}~\cite{yin2025insightvision}, AxiOM~\cite{mazhar2025figurative}, MSAIRS~\cite{shi2025impact}, \benchtestonly{SVBench}~\cite{peng2025svbench}, \benchtestonly{CII-Bench}~\cite{zhang-etal-2025-mllms}, \benchtestonly{VIBE}~\cite{chakraborty-etal-2025-vibe}, \benchtestonly{Hateful Memes}~\cite{kiela2020hatefulmemes}, 
\benchtestonly{PunMemeCN}~\cite{xu-etal-2025-punmemecn}, \benchtestonly{PunchBench}~\cite{ouyang2025punchbench},  
\benchtestonly{CHIME}~\cite{xie-etal-2025-large}, 
\benchtestonly{GOAT-Bench}~\cite{lin2026goat}
\newline
\textbf{[Audio]}  CMSLIU~\cite{li2023discrimination}, \benchtestonly{MMAR}~\cite{NEURIPS2025_610a7d65}, 
\benchtestonly{ParaS2SBench}~\cite{yang2025paras2s}, \benchtestonly{ContraProST}~\cite{li2026plast} 
  \newline
\textbf{[MultiModality]} 
 Genesis~\cite{li2025genesis}  \\

\bottomrule
\end{tabular}
\end{table*}

\subsection{From Basic-Level to High-Level Semantics.}

The evolution from BLSI to HLSI reflects a shift from processing explicit, direct meanings toward interpreting implicit, context-dependent, affective, and socially grounded communication. \textbf{Crucially, mature basic semantic tasks are currently being redefined by HLS requirements.} 

In the text modality, this is driving AI models past literal comprehension and toward pragmatic, creative reasoning that yields sophisticated semantic effects. 
As traditional NLP tasks encounter new bottlenecks when stretched into high-level scenarios, contemporary modeling must pivot to address these implicit complexities directly.
Machine translation is a representative example. 
Although current LLMs have become relatively proficient at basic translation across multiple languages, they still exhibit limitations when handling more semantically complex phenomena, such as metaphors~\cite{wang-etal-2025-drt, wang-etal-2024-mmte}, puns~\cite{ma-etal-2025-pun2pun}, slangs and social media language~\cite{tian2026literalmappingbenchmarkingimproving, guo2025redefining}. 
Paraphrasing tasks have evolved from surface-level reformulation to deeper semantic interpretation. 
For instance, Chinese Idiom Paraphrasing~\cite{qiang2023chinese} depends on implicit knowledge, contextual reasoning and cultural understanding. 
In the field of summarization, the Multimodal Chat Dialogue Summarization Containing Stickers task~\cite{10.1145/3664647.3680978} is proposed to address online chat scenarios where models need to summarize not only textual content but also HLS cues conveyed by sticker images.
Text understanding also moves from literal understanding toward figurative interpretation~\cite{chakrabarty-etal-2022-flute}, pragmatic reasoning~\cite{li2023diplomat}, and other HLS phenomena, such as drivelology~\cite{wang2025drivel}, parody (or fanchuan in Chinese)~\cite{zheng-etal-2025-fanchuan}, buzzwords~\cite{huang-etal-2025-large}, and slang~\cite{mei-etal-2024-slang}. 
From a generative perspective, language models are expected to further enhance their ability to produce creative content involving such complex semantics~\cite{a78fdd6694d54f5491f7393dfc7529e1}, as well as to improve social communicative capabilities such as empathy~\cite{he2025ecc} and persuasion~\cite{jin-etal-2024-persuading}.

In the visual modality, we observe a gradual shift in research focus from the perceptual recognition and generation of simple objects and everyday scenes to more complex visual scenarios involving rich event-level causal logic, deep implicit meanings, and nuanced human emotions. 
Image understanding has evolved from recognizing object attributes and spatial relationships to interpreting latent semantics, metaphorical symbolism, and cultural references. 
Benchmarks such as DEEPEVAL~\cite{yang-etal-2024-large}, II-Bench~\cite{liu2024ii}, CII-Bench~\cite{zhang-etal-2025-mllms}, and InsightVision~\cite{yin2025insightvision} require models to leverage advanced reasoning, extensive knowledge, and cultural nuance to understand deep visual semantics, including philosophical implications, humor, metaphors, and other deep semantics embedded in images. 
Image captioning and visual reasoning have evolved from describing and reasoning about simple scenes to interpreting images with complex semantics. 
CogBench~\cite{song-etal-2025-cognitive} evaluates the cognitive abilities of LVLMs by requiring them to describe the stories depicted in images with dramatic narratives, based on reasoning about events, causal relations, character relationships, and characters’ mental states.
Similarly, R³-VQA~\cite{niu2025r} evaluates the social reasoning abilities of LVLMs in complex video scenarios by requiring models to understand social events, estimate fine-grained mental states such as beliefs, intents, desires, and emotions, and reason over social causal chains. 
Image assessment also moved beyond perceptual quality and visual complexity toward the evaluation of abstract semantic complexity, as exemplified by ISA~\cite{Song_Pang_Tang_Wu_Zhu_2025}, which assesses visual cues, reasoning chains, and causal relations in storytelling images. 
From the perspective of generation tasks, Storytelling Image Generation~\cite{song2025generating} aims to generate logically coherent narratives within a single image through complex Chains-of-Reasoning (CoRs). 
SVBench~\cite{peng2025svbench} moves beyond surface-level perceptual fidelity to evaluate whether video generation models can capture the causal and psychological dynamics underlying socially coherent behavior.

In the audio modality, research has similarly shifted toward uncovering the rich semantic information embedded in audio signals, such as how paralinguistic clues enhance the semantic interpretation of speech. 
MMAR~\cite{NEURIPS2025_610a7d65} evaluates whether audio-language models can move beyond surface-level acoustic and perceptual understanding to perform deep semantic and cultural reasoning over real-world audio scenarios involving sound, music, and speech, where models must connect multiple audio cues through multi-step reasoning to infer higher-level meanings.
Speech understanding has advanced from literal transcription and acoustic feature recognition to the interpretation of beyond-semantic information, emphasizing emotional clues, contextual dynamics, and implicit semantics~\cite{wang2025bossbeyondsemanticspeech}. 
EchoMind~\cite{zhou2025echomindinterrelatedmultilevelbenchmark} further reflects the shift toward high-level audio semantics by evaluating whether Speech Language Models can integrate spoken words with non-lexical vocal cues, infer emotional and contextual meanings, and generate empathetic responses aligned with human communicative intent. 
Speech evaluation has progressed beyond signal quality assessment to measuring paralinguistic awareness and content–style consistency, which further laid foundation for advanced speech interaction. As is shown in ParaS2S~\cite{yang2025paras2s}, speech interaction has advanced from simple command execution to socially intelligent empathetic dialogue by adapting responses to real-time nonverbal cues, such as tone, intonation, and vocalizations.
PLaST~\cite{li2026plast} extends speech translation beyond literal linguistic-content transfer by explicitly modeling paralinguistic cues such as emphasis, emotion, and prosody, which can reshape the semantic interpretation of spoken utterances.
In terms of generation, ParalinGPT~\cite{lin2024paralinguistics} incorporates paralinguistic attributes into spoken dialogue generation, allowing LLMs to produce responses that better reflect the speaker’s sentiment, emotion, and speaking style.

\subsection{High-Level Semantic Tasks}

We summarize the characteristics of HLS understanding and generation tasks from the perspectives of inputs and outputs.

\subsubsection{Semantic Complex Inputs}

As indicated by our definition of HLS and illustrated in Figure~\ref{fig:semantic_up}, HLS is characterized by a high degree of semantic complexity, with inputs that are often semantically dense, contextually extended, cross-modally grounded, and socially situated.

\textit{Interacting Rich Semantic Clues.} 
HLS emerges from the interaction among multiple semantic clues. 
Thus, HLS tasks typically rely on clue-rich inputs, requiring models to reason over these clues for understanding and subsequent generation. 
MOMENTS~\cite{villa-cueva-etal-2025-moments} uses narrative-rich scenarios presented in short films to evaluate the Theory of Mind capabilities of MLLMs. 
VIBE~\cite{chakraborty-etal-2025-vibe} focuses on visual social-pragmatic inference, requiring models to infer emotions, social dynamics, and implicit communicative meanings from non-verbal visual clues in social scenes.
Inspired by the Cookie Theft picture used in human cognitive assessment, CogBench~\cite{song-etal-2025-cognitive} uses storytelling images with rich semantic clues, such as causal relations, character relationships, and mental states, to evaluate the cognitive reasoning abilities of LVLMs.
Benchmarks for implicit meaning understanding in images also evaluate models’ ability to interpret deep semantic meanings through visual clues embedded in images~\cite{liu2024ii, zhang-etal-2025-mllms, yang-etal-2024-large}. 
Metaphor and sarcasm datasets similarly encode tensions between surface expressions and intended meanings. 
Metaphor tasks, such as MUNCH~\cite{tong-etal-2024-metaphor}, require models to transcend lexical similarity to capture underlying relational structures. 
Sarcasm tasks like iSarcasmEval~\cite{abu-farha-etal-2022-semeval} center on identifying the discrepancy between literal and intended meanings. 
The FLUTE~\cite{chakrabarty-etal-2022-flute} task extends this challenge to explanatory reasoning to justify the contradiction between premises and hypotheses. 
Empathy and persuasion inputs also combine events, emotions, intentions, goals, and strategies. EmotionQueen~\cite{chen-etal-2024-emotionqueen} assesses emotional intelligence by challenging models to decipher the interaction between explicit situational events and latent emotional states. 
SafePersuasion~\cite{kong-etal-2025-safepersuasion} requires models to differentiate between rational persuasion and manipulation by analyzing the complex interaction of subtle linguistic styles and cognitive heuristics. 
In addition, benchmarks based on memes~\cite{kiela2020hatefulmemes}, puns~\cite{xu-etal-2024-good}, and related phenomena are also characterized by rich semantic clues. 
PUNMEMECN~\cite{xu-etal-2025-punmemecn} evaluates models’ ability to understand puns in memes. 

\textit{Extended Contextual Scope.} 
The semantic complexity of HLS also arises from their extended contextual scope. 
The input unit in HLS tasks often goes beyond a local sentence, image, or dialogue turn, requiring models to integrate information across longer contexts, multiple events, broader narratives, or evolving interaction histories. 
For humor, STRIP-CIPHER~\cite{wang2025beyond} and PixelHumor~\cite{ryan-etal-2025-humor} use sequences of comic panels whose meanings depend on temporal and causal continuity. 
Sarcasm tasks such as DocMSU embed sparse sarcastic clues in long news documents and paired images, requiring integration of distant contextual signals~\cite{Du_Nan_Zhang_Xie_Xu_Fan_Cui_Tao_Jiang_2024}. Empathy-related settings like DiPlomat situate pragmatic meaning in multi-turn conversations, where interpretation depends on discourse history~\cite{li2023diplomat}. 
In persuasion, PersuasiveToM~\cite{yu2025persuasivetom} and NegotiationToM~\cite{chan2024negotiationtom} further extend this structure by treating belief, preference, and intention tracking across interaction turns as part of the input. Narrative tasks such as NarrativeQA~\cite{kocisky-etal-2018-narrativeqa} further reflect extended-context understanding, requiring models to integrate dispersed events, character trajectories, and plot relations across stories or novels. Across these tasks, meaning emerges from extended temporal, discourse, and interactional context rather than isolated inputs. 

\textit{Cross-Modal Semantic Grounding.}
HLS is also grounded jointly in language, vision, speech, and embodied behavior. 
Humor-related tasks such as PunchBench~\cite{ouyang2025punchbench}, YesBut~\cite{nandy-etal-2024-yesbut}, and InsightVision~\cite{yin2025insightvision} require models to infer punchlines, satire, or implicit meanings from image-text relations, symbols, and background knowledge. 
V-AlphaSocial~\cite{lin-etal-2025-v} evaluates whether LVLMs can integrate textual and visual cues for social commonsense reasoning. 
PunchBench further verifies that models are genuinely reasoning through multimodal interactions rather than relying on linguistic shortcuts~\cite{ouyang2025punchbench}. 
Metaphor understanding in MultiCMET~\cite{zhang-etal-2023-multicmet} and V-FLUTE~\cite{saakyan-etal-2025-understanding} distributes figurative meanings across images and captions, requiring alignment between visual and linguistic signals. 
Empathy tasks such as MEDIC~\cite{zhu2023medic} model counseling signals through conversational content, facial behavior, body movement, and speech, while EchoMind~\cite{zhou2025echomindinterrelatedmultilevelbenchmark} shows that vocal delivery can alter emotional meaning. 
These examples illustrate that HLS arises from the interaction of multiple modalities rather than a single source.

\textit{Socially and Culturally Situated Signals.}
Many HLS phenomena involve clues whose interpretation depends on social norms, cultural conventions, and audience background. 
In humor, Chumor~\cite{he-etal-2025-chumor} and PunMemeCN~\cite{xu-etal-2025-punmemecn} rely on Chinese homophones, character forms, cultural references, and meme conventions. 
Sarcasm perception in MultiPICo~\cite{casola-etal-2024-multipico} varies across languages, regions, and demographic groups. 
Empathy-related datasets such as EmpathyFromPerspectives~\cite{chen-etal-2025-empathy} and EmpathicStories++~\cite{shen-etal-2024-empathicstories} show that personal perspectives and prior experiences shape emotional interpretation. 
In persuasion, PVP~\cite{kim-etal-2025-pvp} demonstrates that effectiveness depends on viewer demographics and psychological traits, while SOCKET~\cite{choi-etal-2023-llms} and V-SOCIAL~\cite{lin-etal-2025-v} evaluate social norms and commonsense grounded in linguistic and visual situations. 
Together, these tasks highlight that meaning is deeply conditioned by social context, cultural knowledge, and audience-specific factors.

\subsubsection{Interpretive and Explanatory Outputs in HLS Understanding}

The complexity of HLS inputs drives understanding tasks beyond explicit recognition and literal answering, requiring models to uncover implied meanings and explain how semantic effects are produced.

\textit{Nonliteral Meaning Identification.}
A central form of HLS output is the identification of meanings that are implied rather than directly expressed. 
For humor, the required output is often the punchline or incongruous relation implied by multiple cues; in PunchBench, this is instantiated as identifying the humorous implication created by an image--caption pair~\cite{ouyang2025punchbench}. 
For metaphor, the output centers on selecting the intended nonliteral interpretation: Fig-QA uses the intended implication of a figurative expression as the answer~\cite{liu-etal-2022-testing}, and MUNCH formulates metaphor understanding as selecting apt paraphrases over lexically related but semantically inapt alternatives~\cite{tong-etal-2024-metaphor}. 
For sarcasm, the output target is the speaker's intended stance behind the literal utterance; iSarcasmEval asks models to identify sarcastic given a sarcastic sentence and its non-sarcastic rephrase~\cite{abu-farha-etal-2022-semeval}, while DocMSU defines document-level multimodal sarcasm understanding as detecting and localizing sarcasm from textual and visual evidence~\cite{Du_Nan_Zhang_Xie_Xu_Fan_Cui_Tao_Jiang_2024}. 
Empathy and persuasion extend this output form to social and affective interpretation: EmotionQueen formulates empathetic understanding as recognizing key events, mixed events, implicit emotions, and user intentions, followed by empathetic responses~\cite{chen-etal-2024-emotionqueen}, and PersuasiveToM defines dialogue-level ToM tasks over beliefs, desires, intentions, and persuasive strategies~\cite{yu2025persuasivetom}. 
More general benchmarks such as DiPlomat~\cite{li2023diplomat} and InsightVision~\cite{yin2025insightvision} further define outputs around implicit pragmatic meanings and multilevel visual implications. 
Together, these datasets show that implicitness is central to HLS: the expected output specifies an interpretation that is cued by the input but not directly stated.

\textit{Mechanism-Aware Explanation.}
Beyond identifying implicit meanings, another HLS understanding output form is mechanism-aware explanation, where the answer accounts for how an intended semantic effect is produced. 
For humor, explanations usually identify the incongruity, punchline, or background knowledge that makes the content funny. 
The New Yorker Caption Contest~\cite{hessel-etal-2023-androids} asks for explanations of why a caption is funny in relation to a given cartoon, Chumor~\cite{he-etal-2025-chumor} focuses on Chinese joke explanation with linguistic and culture-specific knowledge, and Comparing Apples to Oranges~\cite{loakman-etal-2025-comparing} provides reference explanations across diverse joke types. 
For sarcasm, explanations focus on the conflict between surface expression and intended meaning. 
WITS~\cite{kumar-etal-2022-become} asks for explanations of sarcastic connotations in multimodal dialogues, and YesBut~\cite{nandy-etal-2024-yesbut} focuses on the conflicting visual scenario that produces a satirical effect. 
Narrative understanding introduces another form of explanatory reasoning, as FLAWEDFICTIONS~\cite{ahuja2025finding} and DetectiveQA~\cite{xu2025detectiveqaevaluatinglongcontextreasoning} require models to reconstruct how events, motivations, clues, and causal relations jointly form a coherent story world.
More general figurative-semantics tasks extend explanation beyond a single phenomenon: FLUTE formulates figurative understanding as textual entailment with explanations, and V-FLUTE extends this setting to vision-language figurative entailment~\cite{chakrabarty-etal-2022-flute,saakyan-etal-2025-understanding}. 
A similar explanatory form appears in GODBench's Video Comment Art task: given a highly upvoted ``GOD-level'' comment for a video, the explanation output identifies its Comment Art dimensions and accounts for why the comment is creative or engaging~\cite{lei-etal-2025-godbench}. 
Together, these tasks show that mechanism-aware explanation turns HLS outputs from identifying an implied meaning to accounting for the semantic mechanism that makes the effect interpretable.

\subsubsection{Outputs for Semantic Effect Construction in HLS Generation} 

HLS generation tasks typically require models to first develop a comprehensive understanding of the complex input, and then generate outputs by leveraging cognitive reasoning, external knowledge, and creativity, thereby enabling the generated content and the input to jointly form higher-level semantic effects.
As mentioned earlier, the input may contain dialogue history, visual content, emotional states, persuasive goals, figurative prompts, or multimodal clues, so the model must infer implicit relations such as intent, attitude, affect, incongruity, or source-target mapping before generating. 
The generation task is therefore more complex than ordinary conditional generation: the output must not only remain fluent and relevant to the input, but also externalize this inferred meaning through text, image, speech, or multimodal content. 
In this sense, HLS generation links input understanding with output realization, requiring models to generate outputs that, together with the input, produce the intended HLS effects. 

Across topics, this output richness is often produced by coordinating multiple semantic conditions into a generated effect. 
In humor generation, conversational context, speaker role, or visual content is paired with an unexpected but interpretable response to realize incongruity-based amusement. 
TalkFunny~\cite{Chen_Yuan_Liu_Liu_Guan_Guo_Peng_Liu_Li_Xiao_2024} studies humorous dialogue response generation from context, C3~\cite{li-etal-2023-language} formulates Chinese comical crosstalk as role-based multi-turn script generation, and Oogiri-GO~\cite{a78fdd6694d54f5491f7393dfc7529e1} combines image understanding with textual reframing to generate vision-language humor. 
In metaphor generation, the output must jointly express a source domain and a target meaning so that an implicit mapping becomes perceptible; HAIVMet~\cite{chakrabarty-etal-2023-spy} exemplifies this requirement by generating images that visually convey the meaning of linguistic metaphors. 
In sarcasm generation, semantic richness comes from aligning a surface expression with an incongruent evaluative stance, as in MuSG's image-grounded sarcastic description generation~\cite{zhao-etal-2023-multi}. 
In empathy generation, the model must combine the user's affective state with suitable supportive content and modality: ECC~\cite{he2025ecc} uses emotion-cause relations to guide empathetic responses, while OSUM-EChat~\cite{geng2025osum} extends this requirement to spoken dialogue where vocal delivery also contributes to the empathetic effect. 
In persuasion generation, audience information, communicative intent, and strategy selection jointly shape outputs toward a desired social outcome, as shown by DailyPersuasion's strategy-aware persuasive dialogue generation and PVP's personalized visual persuasion setting~\cite{jin-etal-2024-persuading,kim-etal-2025-pvp}. 
More general creative generation tasks show a similar pattern in wordplay, where keywords and lexical ambiguity are coordinated to produce humorous double meanings, as in ExPUNations~\cite{sun-etal-2022-expunations}. 
Overall, these studies suggest that semantically rich generation is defined by the model's ability to integrate contextual, modal, affective, strategic, and figurative constraints into an intended HLS effect.

%% file: 04_1_high-level_semantic_intelligence_und.tex
\section{High-Level Semantic Understanding}
\label{sec:hlsu}

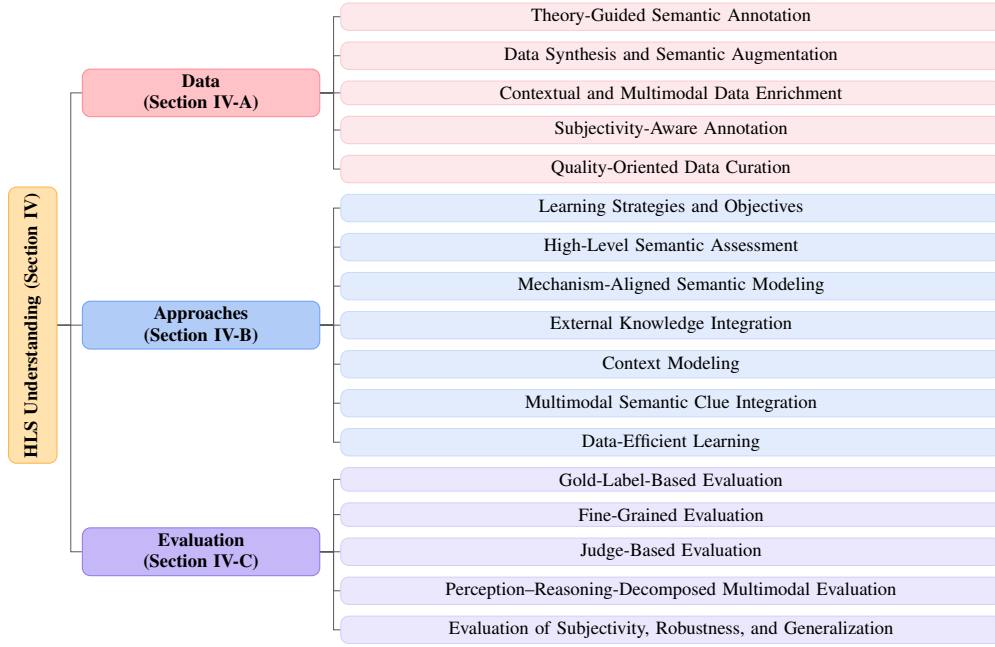
\begin{figure*}[!t]
  \centering

  \definecolor{HLSRootFill}{RGB}{255,226,174}
  \definecolor{HLSRootDraw}{RGB}{255,183,77}

  \definecolor{DataHeadFill}{RGB}{255,194,199}
  \definecolor{DataHeadDraw}{RGB}{255,140,150}
  \definecolor{DataLeafFill}{RGB}{253,233,235}
  \definecolor{DataLeafDraw}{RGB}{248,205,210}

  \definecolor{ApproachHeadFill}{RGB}{178,204,248}
  \definecolor{ApproachHeadDraw}{RGB}{112,163,246}
  \definecolor{ApproachLeafFill}{RGB}{226,236,252}
  \definecolor{ApproachLeafDraw}{RGB}{203,220,249}

  \definecolor{EvalHeadFill}{RGB}{198,184,248}
  \definecolor{EvalHeadDraw}{RGB}{145,124,226}
  \definecolor{EvalLeafFill}{RGB}{235,231,252}
  \definecolor{EvalLeafDraw}{RGB}{217,208,247}

  \tikzset{
    root node/.style={
      draw=HLSRootDraw,
      fill=HLSRootFill,
      align=center,
      thin,
      rounded corners=3,
    },
    data head/.style={
      draw=DataHeadDraw,
      fill=DataHeadFill,
      align=center,
      thin,
      text width=3cm,
      rounded corners=3,
    },
    data leaf/.style={
      draw=DataLeafDraw,
      fill=DataLeafFill,
      align=center,
      thin,
      text width=8.6cm,
      rounded corners=3,
    },
    approach head/.style={
      draw=ApproachHeadDraw,
      fill=ApproachHeadFill,
      align=center,
      thin,
      text width=3cm,
      rounded corners=3,
    },
    approach leaf/.style={
      draw=ApproachLeafDraw,
      fill=ApproachLeafFill,
      align=center,
      thin,
      text width=8.6cm,
      rounded corners=3,
    },
    evaluation head/.style={
      draw=EvalHeadDraw,
      fill=EvalHeadFill,
      align=center,
      thin,
      text width=3cm,
      rounded corners=3,
    },
    evaluation leaf/.style={
      draw=EvalLeafDraw,
      fill=EvalLeafFill,
      align=center,
      thin,
      text width=8.6cm,
      rounded corners=3,
    }
  }

  \begin{forest}
    nonleaf/.style={
      font=\bfseries\scriptsize
    },
    for tree={
      font=\scriptsize,
      text=black,
      anchor=west,
      inner sep=2pt,
      l sep=7.5pt,
      s sep=4pt,
      fit=tight,
      grow'=east,
      edge={
        ultra thin,
        draw=black!65
      },
      parent anchor=east,
      child anchor=west,
      if n children=0{}{
        nonleaf
      },
      edge path={
        \noexpand\path[\forestoption{edge}]
        (!u.parent anchor) -- +(5pt,0) |- (.child anchor)
        \forestoption{edge label};
      },
      if={isodd(n_children())}{
        for children={
          if={equal(n,(n_children("!u")+1)/2)}{
            calign with current
          }{}
        }
      }{}
    }
    [
      {\rotatebox{90}{HLS Understanding (Section~\ref{sec:hlsu})}},
      root node,
      text width=0.5cm,
      font=\bfseries\scriptsize,
      inner sep=2pt,
      minimum height=1.9cm,
      minimum width=0.45cm
      [
        Data \\(Section~\ref{sec:hlsu_data}),
        data head,
        for descendants={data leaf}
        [
          Theory-Guided Semantic Annotation
        ]
        [
          Data Synthesis and Semantic Augmentation
        ]
        [
          Contextual and Multimodal Data Enrichment
        ]
        [
          Subjectivity-Aware Annotation
        ]
        [
          Quality-Oriented Data Curation
        ]
      ]
      [
        Approaches \\(Section~\ref{sec:hlsu_approach}),
        approach head,
        for descendants={approach leaf}
        [
          Learning Strategies and Objectives
        ]
        [
          High-Level Semantic Assessment
        ]
        [
          Mechanism-Aligned Semantic Modeling
        ]
        [
          External Knowledge Integration
        ]
        [
          Context Modeling
        ]
        [
          Multimodal Semantic Clue Integration
        ]
        [
          Data-Efficient Learning
        ]
      ]
      [
        Evaluation \\(Section~\ref{sec:hlsu_eval}),
        evaluation head,
        for descendants={evaluation leaf}
        [
          Gold-Label-Based Evaluation
        ]
        [
          Fine-Grained Evaluation
        ]
        [
          Judge-Based Evaluation
        ]
        [
          Perception--Reasoning-Decomposed Multimodal Evaluation
        ]
        [
          {Evaluation of Subjectivity, Robustness, and Generalization}
        ]
      ]
    ]
  \end{forest}

  \caption{
    Taxonomy of High-Level Semantic (HLS) understanding methods, covering
    data construction, modeling approaches, and evaluation protocols.
  }
  \label{fig:understanding-taxonomy}
\end{figure*}

HLS understanding concerns meanings that are not directly manifested by surface observations. 
Unlike BLS understanding, where target meanings are typically grounded in observable entities, actions, attributes, or factual relations, HLS understanding requires models to infer latent, context-dependent meanings from interrelated semantic clues. 
Therefore, the key object of understanding is often not an isolated token, image region, acoustic feature, or dialogue turn, but the structured relations among clues that jointly give rise to high-level meaning.

\subsection{Data}
\label{sec:hlsu_data}

Data design plays a central role in HLS understanding because it determines the semantic distinctions, contextual evidence, and human perspectives available to models. Existing work improves HLS data along five complementary dimensions: enriching annotations with theory-guided semantic structures, expanding data coverage through synthesis and augmentation, incorporating contextual and multimodal evidence, preserving subjective variation, and strengthening data quality through targeted curation.

\subsubsection{Theory-Guided Semantic Annotation}

Rather than relying solely on coarse task labels, HLS datasets increasingly use linguistic, cognitive, and social theories to annotate the structures and mechanisms underlying semantic phenomena. For figurative language, ExPUN supplements pun instances with explanations and ratings of funniness, understandability, and offensiveness~\cite{sun-etal-2022-expunations}. A Chinese metaphor dataset represents metaphor identification as span-based relation extraction by annotating source and target spans and their relation types~\cite{chen-etal-2023-chinese}, while FLUTE constructs natural-language inference pairs with explanations for sarcasm, simile, metaphor, and idiom~\cite{chakrabarty-etal-2022-flute}. For narrative understanding, StorySeeker operationalizes narratological definitions through an expert-developed codebook, providing document-level story labels together with story and event spans for narratives embedded in broader online discourse~\cite{antoniak-etal-2024-people}.

Theory-guided annotation also captures social, pragmatic, and cross-modal mechanisms. ALOE models conversational empathy through appraisal dimensions and Target--Observer alignment, emphasizing perspective-taking rather than surface emotional similarity~\cite{yang-jurgens-2024-modeling}. SafePersuasion distinguishes rational persuasion from manipulation and provides fine-grained strategy labels~\cite{kong-etal-2025-safepersuasion}. For multimodal sarcasm, MSD-CoT supplies structured reasoning chains covering visual description, textual meaning, sentiment, cross-modal incongruity, and final inference~\cite{InterARM2026}. Together, these datasets transform HLS annotations from opaque outcomes into structured representations of semantic composition, reasoning processes, and communicative mechanisms.

\subsubsection{Data Synthesis and Semantic Augmentation}

Because high-quality HLS data are costly to collect, recent studies use LLMs and automatic augmentation to generate scenarios, contrastive examples, explanations, and semantic distractors, typically followed by human review or task-specific filtering. 
EmotionQueen uses GPT-4 to generate statements across diverse life situations for recognizing key events, mixed events, implicit emotions, and intentions~\cite{chen-etal-2024-emotionqueen}. 
Other methods construct contrastive or counterfactual pairs: humorous texts can be transformed into aligned non-humorous counterparts~\cite{horvitz-etal-2024-getting}, while RedSD rewrites sarcastic dialogues into context-preserving non-sarcastic variants and validates them for label flipping, plausibility, contextual adequacy, and content preservation~\cite{hong-etal-2025-rhetorical}. 
Controlled augmentation also supports semantic reasoning. 
MUNCH constructs apt and inapt metaphor paraphrases using WordNet synonyms and hypernyms as controlled distractors~\cite{tong-etal-2024-metaphor}, whereas PersuasiveToM generates multiple-choice questions and semantic distractors for reasoning about beliefs, desires, intentions, and persuasive strategies~\cite{yu2025persuasivetom}. 
For multimodal narrative reasoning under imperfect information, Zhong et al. coordinate generation and critic agents to synthesize murder-mystery scripts containing role backgrounds, textual and visual clues, interactive dialogues, and grounded multi-hop reasoning traces~\cite{zhong-etal-2026-collaborative}. 
Together, these approaches improve scalability while preserving semantic control through contrastive construction, structured synthesis, and explicit validation.

\subsubsection{Contextual and Multimodal Data Enrichment}

Since HLS often emerges from relations across modalities or interaction contexts, recent datasets enrich isolated inputs with visual, acoustic, conversational, or structured semantic information. SarcNet separately annotates textual, visual, and multimodal sarcasm, revealing cases in which sarcasm specifically arises from cross-modal contrast~\cite{yue-etal-2024-sarcnet}. HumorDB provides minimally contrastive funny and non-funny image pairs created through editing or inpainting; by controlling irrelevant visual differences, the dataset tests whether models identify humor-producing elements~\cite{jain2025humordb}. CM3D annotates source- and target-domain elements in Chinese multimodal advertisements, making cross-modal metaphor mappings explicit~\cite{zhang2025towards}. Contextual enrichment further extends to interaction-level signals: MEDIC provides therapeutic dialogues with turn-level textual, acoustic, and visual features annotated for expressions of experience, emotional reactions, and cognitive reactions~\cite{zhu2023medic}. Collectively, these datasets enable more diagnostic evaluation through modality-specific labels, controlled contrasts, structured relations, and interaction histories.

\subsubsection{Subjectivity-Aware Annotation}

Because HLS judgments vary across individuals and communities, subjectivity-aware datasets represent annotations as preferences, distributions, or perspective-dependent scores rather than enforcing a single ground-truth label. In humor, Murakami et al.\ cluster users based on Oogiri voting logs and apply Bradley--Terry--Luce models to estimate group-specific weights for humor factors~\cite{murakami2026wholaughs}. Hessel et al.\ derive caption matching, ranking, and explanation tasks from reader and editor preferences in the \textit{New Yorker} Caption Contest, separating visual understanding and caption appropriateness from perceived humor quality~\cite{hessel-etal-2023-androids}. Other datasets preserve user- or annotator-dependent perceptions more directly. SENSE-7 collects turn-level user judgments along seven dimensions of perceived empathy in sustained conversations~\cite{suh2026sense}; MultiPICo provides multilingual, disaggregated irony annotations with annotator metadata, retaining demographic and cultural variation~\cite{casola-etal-2024-multipico}; and PVP associates visual-persuasion judgments with viewers' traits, values, and moral foundations~\cite{kim-etal-2025-pvp}. By retaining these differences, such datasets treat disagreement as meaningful variation rather than annotation noise.

\subsubsection{Quality-Oriented Data Curation}

Quality-oriented curation addresses three recurring risks in HLS data: invalid semantic interpretations, unreliable supervision, and benchmark artifacts. Annotation validation ensures that labels and explanations capture the intended meaning. Chumor manually verifies whether LLM-generated explanations fully account for Chinese jokes~\cite{he-etal-2025-chumor}, while SafePersuasion employs iterative expert annotation, agreement tracking, and the removal of ambiguous instances~\cite{kong-etal-2025-safepersuasion}. Supervision noise can also be explicitly modeled or filtered: UPLME represents noisy empathy scores through heteroscedastic uncertainty~\cite{hasan2025uplme}, whereas CONELA combines training dynamics with human agreement to identify low-quality implicit-hate instances~\cite{kim-etal-2025-analyzing}. Finally, adversarial and counterfactual variants can expose benchmark shortcuts. Literal paraphrases of metaphorical inputs, for example, reveal model sensitivity to lexical overlap and sentence length~\cite{sanchez-bayona-agerri-2025-metaphor}. These practices frame HLS data quality as the joint problem of ensuring semantic validity, supervision reliability, and diagnostic robustness.

\subsection{Approaches}
\label{sec:hlsu_approach}

\subsubsection{Learning Strategies and Objectives.}

We categorize learning strategies for HLS understanding according to the structure of their supervision signals: direct supervision with task-specific semantic labels, objective shaping with auxiliary semantic constraints, and reasoning- or feedback-guided optimization. These strategies are not mutually exclusive; rather, they reflect increasing efforts to move beyond surface label prediction toward semantically grounded and interpretable understanding.

\textbf{Supervised learning with semantic labels} remains the dominant paradigm. It formulates HLS understanding as classification, regression, ranking, extraction, or sequence labeling and typically adopts standard objectives such as cross-entropy, mean squared error, ranking loss, or instruction-finetuning likelihood. The form and granularity of supervision vary across tasks. Humor understanding, for example, ranges from utterance-level detection and intensity regression to word-level laughter prediction~\cite{alnajjar-etal-2022-laugh,barriere-etal-2025-standup4ai}. Empathy models employ multitask or multi-label supervision to jointly capture empathic intent and perceived dimensions~\cite{jiang2023empathyintent,xu2024empeval}. Metaphor understanding has been modeled through both classification augmented with linguistically motivated signals, such as MIP and SPV, and structured extraction of source--target spans and their relations~\cite{zhang2022misnet,chen-etal-2023-chinese}. Similar formulations have been applied to persuasion, where rhetorical modes such as logos, pathos, and ethos provide task-specific semantic labels~\cite{alyahya2025hatred}. Although direct supervision offers stable and controllable learning signals, its effectiveness depends heavily on annotation quality and granularity, and it remains vulnerable to class imbalance, domain shift, and superficial shortcuts.

\textbf{Objective shaping with auxiliary supervision} enriches direct task objectives with constraints designed to capture semantic relations that cannot be adequately expressed by target labels alone. One line of work introduces structured semantic objectives. For instance, metaphor models employ attribute mapping, domain contrast, or relation-based ranking to capture source--target correspondences and domain incongruity~\cite{tian-etal-2023-modeling,sengupta-etal-2023-modeling}. Another line explicitly addresses subjective or noisy supervision through uncertainty estimation, semantic-similarity alignment, and outlier masking~\cite{hasan2025uplme,25EnhancingSemanticAwareness}. Contrastive and counterfactual learning further encourage models to distinguish semantically diagnostic variations: minimally edited humorous and non-humorous pairs reduce lexical shortcuts~\cite{horvitz-etal-2024-getting}, while curriculum learning and hard-example selection improve figurative-language and implicit-hate recognition~\cite{zhou2023clcl,jiang-2025-learn}. Similar contrastive objectives have also been applied to imbalanced multimodal persuasion recognition~\cite{cui2023multimodal}. Collectively, these methods shift learning from reproducing annotations to modeling the contrasts, relations, and uncertainties underlying them. Their gains, however, depend on whether the auxiliary, contrastive, or synthetic signals faithfully represent the target semantics.

\textbf{Reasoning- and feedback-guided optimization} extends the optimization target from final predictions to the processes through which semantic conclusions are formed. At training time, structured rationales and preferences can be incorporated through SFT, DPO, PPO, or GRPO. In humor understanding, pairwise preference learning supports comparative evaluation and can be less noisy than pointwise scoring~\cite{zhou2025humoralignment,wang2025innovative}. For sarcasm and emotion recognition, structured reasoning dimensions and task-specific rewards are used to optimize both prediction accuracy and reasoning quality~\cite{InterARM2026,yang2025sarcasmr1,song2025emotion}. 
For multimodal narrative reasoning, Zhong et al. combine LoRA-based SFT on structured reasoning traces with GRPO guided by agent-monitored rewards for correctness, role consistency, and image--clue grounding~\cite{zhong-etal-2026-collaborative}. 
Reasoning traces generated by multimodal models can also be distilled into smaller supervised classifiers, as demonstrated in multimodal metaphor detection~\cite{xu2024c4mmd}. At inference time, models may instead be guided without parameter updates through iterative critique, feedback memory, explicit rubrics, or adaptive demonstrations~\cite{liu2026florence,suh2026sense}. This family reflects a shift from optimizing final predictions alone to jointly optimizing predictions and the reasoning processes that support them. Such methods are particularly relevant to subjective and pragmatic HLS tasks. Nevertheless, improved rationales or preference scores do not necessarily indicate better semantic understanding, making it essential to evaluate the faithfulness of the reasoning process itself.

Overall, these three families provide complementary forms of supervision. Direct semantic labels establish task-level decision boundaries, auxiliary objectives encode finer semantic structures and robustness constraints, and reasoning- or feedback-guided methods regulate how models arrive at their predictions. Their combination represents a promising direction for developing HLS understanding systems that are both accurate and semantically faithful.

\subsubsection{High-Level Semantic Assessment}

HLS assessment aims to measure properties of HLS, such as its complexity, richness, and intensity, as well as the effects it produces. Such assessment is valuable for diagnosing models' semantic understanding, evaluating HLS tasks, and supporting controllable generation. However, the abstract, context-dependent, and subjective nature of HLS makes reliable measurement difficult. Existing approaches can be broadly divided into intrinsic assessment of semantic content and effect-based assessment through psychological or behavioral responses.

One line of work \textbf{quantifies intrinsic semantic attributes}, including conceptual magnitude, linguistic complexity, and visual semantic richness. Semantic size captures the perceived magnitude or salience of concrete and abstract concepts~\cite{scott2019glasgow}. Recent work finds that multimodal LLMs align more closely with human judgments of semantic size than text-only models, indicating the importance of perceptual grounding~\cite{yao2025deep}. For linguistic complexity, RepublicQA considers conceptual difficulty, lexical diversity, and structural contrast~\cite{zhang2026semantic}, while studies of spoken language understanding combine lexical statistics with embedding-based geometric measures and show that greater complexity is often associated with poorer model performance~\cite{mckenna2020semantic}. In the visual modality, most benchmarks focus on perceptual image complexity, which is more closely related to BLS~\cite{feng2023ic9600}. By contrast, VLISA assesses the semantic richness of storytelling images by converting entities, clues, character relations, events, causal links, and mental states into language-based CoR features for predicting entity- and semantic-complexity scores~\cite{Song_Pang_Tang_Wu_Zhu_2025}. These studies extend intrinsic assessment from surface complexity toward the conceptual and relational structures underlying HLS.

Another line of work \textbf{assesses semantic effects through psychological responses and feedback signals}. The Psychological Depth Scale evaluates creative stories along emotional provocation, empathy, engagement, authenticity, and narrative complexity, combining human ratings with LLM-as-a-judge evaluation~\cite{harel-canada-etal-2024-measuring}. Empathy studies similarly employ psychological scales, author self-reports, interlocutor ratings, third-party annotations, and continuous scores to measure dimensions such as empathic concern, personal distress, emotional reaction, interpretation, and exploration~\cite{buechel2018modeling,ma2025emprl,omitaomu2022empathic,hasan-etal-2024-llm}. Other methods use observable or model-based proxy signals: sitcom laughter indicates humor intensity~\cite{alnajjar-etal-2022-laugh}, language-model surprise approximates expectation violations in humor and metaphor~\cite{bunescu-uduehi-2022-distribution}, and changes in audience votes before and after debate speeches measure persuasive impact~\cite{bai2021m2p2}. 
Short-drama assessment further aggregates consumption, completion, interaction, and search behaviors into a user-centric indicator of visual and narrative quality~\cite{liu2026bridging}. 
Although these signals capture semantic effects in realistic settings, they may also be influenced by audience characteristics and other contextual factors.

Overall, HLS assessment is moving beyond surface lexical and perceptual complexity toward more abstract and human-centered indicators. Intrinsic measures characterize the conceptual and relational properties of semantic content, whereas effect-based measures capture its psychological reception and interaction outcomes.

\subsubsection{Mechanism-Aligned Semantic Modeling}

High-level meanings often arise not directly from surface expressions, but from latent mechanisms underlying semantic content and the social situations in which expressions are produced and interpreted. Mechanism-aligned semantic modeling therefore seeks to reconstruct the processes through which observable clues give rise to implicit meanings. Existing methods can be broadly organized according to where the underlying semantic mechanism is located: within the semantic structure of the content or within the social and pragmatic interaction surrounding it.

The first family performs \textbf{content-internal mechanism reconstruction} by explicitly modeling the gap between literal expressions and their intended or inferred meanings. A common approach is to represent semantic incongruity or violated expectations. In sarcasm understanding, DIP models factual and affective incongruity through semantic-distribution and sentiment-contrast modules, while ITFNet learns fact- and sentiment-based tension to characterize context-dependent incongruity~\cite{wen2023dip,zhang-etal-2025-incongruity}. Humor research similarly models the mechanisms connecting context to a punchline: PunchBench evaluates whether MLLMs can reason over visual--textual contrast or alignment~\cite{ouyang2025punchbench}, whereas TalkFunny annotates intermediate structures involving implicit entities, commonsense relations, and humorous incongruity~\cite{Chen_Yuan_Liu_Liu_Guan_Guo_Peng_Liu_Li_Xiao_2024}. For metaphor, MisNet and BasicBERT follow MIP-style reasoning by separating basic meaning from contextual meaning and modeling their incongruity through SPV-related signals~\cite{zhang2022misnet,li2023basicmeaning}. AIDIL and Sengupta et al.\ expose richer source--target structures by modeling domain inconsistency, shared attributes, source domains, and highlighted target aspects~\cite{tian-etal-2023-modeling,sengupta-etal-2023-modeling}. Related principles extend to broader figurative language: $(RSA)^2$ treats rhetorical strategy as a latent variable within the Rational Speech Act framework~\cite{spinoso-di-piano-etal-2025-rsa}, while MSCLM models idioms through literal--idiomatic and cross-context inconsistencies~\cite{Wu_Hu_Zhang_Zhi_Su_Sha_2024}. Across these tasks, the shared objective is to replace direct label prediction with intermediate representations of contrasts, mappings, expectations, or pragmatic strategies that explain how non-literal meanings are derived.

The second family performs \textbf{social-pragmatic mechanism modeling}, reconstructing how communicative goals, speaker and audience states, interpersonal relations, and interaction history shape the interpretation and effects of HLS. This approach is particularly relevant to empathy, persuasion, and humor, whose meanings depend on the relationship between an expression, its producer, and its audience. In empathy understanding, communicative intents and perceived empathy dimensions are modeled jointly, reflecting that empathy depends on speaker purpose and listener perception rather than emotional wording alone~\cite{xu2024empeval,jiang2023empathyintent}. Appraisal-based modeling further characterizes empathy as alignment between a target's appraisal and an observer's response~\cite{yang-jurgens-2024-modeling}. Persuasion methods similarly incorporate rhetorical and audience-related mechanisms: Aristotle's logos, pathos, and ethos provide interpretable categories for persuasive intent~\cite{alyahya2025hatred}, while Theory-of-Mind-based approaches reconstruct audience states. ToMAP predicts counterclaims and opponent attitudes before selecting persuasive strategies~\cite{han2025tomap}, and ToM-Agent tracks beliefs, desires, and intentions throughout a conversation and updates them through counterfactual reflection~\cite{yang2025large}. In humor, social-pragmatic mechanisms are captured through preference learning and personalization: preference-based training aligns multimodal humor evaluation with human rankings~\cite{zhou2025humoralignment}, whereas CIHR combines general humor mechanisms with user profiles to account for individual differences~\cite{zhu2025cihr}. These methods thus explain high-level meaning through the interaction among semantic content, communicative participants, and the surrounding social context.

Overall, the two families locate semantic mechanisms at different but complementary levels. Content-internal mechanism reconstruction explains how implicit meanings emerge from incongruity, mapping, and latent structures within semantic content, whereas social-pragmatic mechanism modeling explains how such meanings are shaped by communicative goals, mental states, audience preferences, and interaction contexts. Together, they suggest that HLS understanding requires not only identifying high-level meanings, but also representing the mechanisms that make those meanings interpretable in both semantic content and social interaction.

\subsubsection{External Knowledge Integration}

Because HLS often depends on information that is not explicitly expressed in the input, a central challenge is how to incorporate external knowledge into models for reliable semantic reasoning. 
Existing methods can be broadly divided according to how knowledge is introduced: retrieval-augmented grounding, integration of structured symbolic resources, and parametric knowledge fusion.

\textbf{Retrieval-augmented knowledge grounding} provides non-parametric access to evidence that may not be reliably stored in the base model. Methods in this family mainly differ in their retrieval sources, knowledge granularity, and mechanisms for incorporating retrieved evidence into reasoning. FLoReNce retrieves prior reasoning traces and judge critiques from a feedback memory to guide subtle humor reasoning while keeping the backbone model frozen~\cite{liu2026florence}. ImaRA constructs and retrieves theory-grounded imaginative frames involving source and target domains, domain inconsistency, and attribute similarity to support low-resource metaphor detection and explanation~\cite{tian-etal-2025-imara}. For target-dependent semantic tasks, MSME retrieves and filters background evidence about unseen targets for zero-shot stance detection~\cite{zhang2026msme}, while LAD transforms visual clues into textual abstractions and retrieves out-of-domain contextual knowledge to resolve image metaphors with missing context~\cite{zhang2025let}. Retrieval offers controllable and inspectable grounding, but its effectiveness depends on evidence quality, relevance filtering, and the model's ability to reconcile retrieved knowledge with its internal priors.

A second family \textbf{integrates structured symbolic resources and knowledge graphs} into representations, graph structures, or prompts. Such resources externalize conceptual attributes, event consequences, commonsense relations, and coded meanings that are difficult to infer from surface observations alone. In metaphor understanding, AIDIL retrieves source- and target-concept attributes from ConceptNet and models their similarity and domain inconsistency~\cite{tian-etal-2023-modeling}, whereas WPDM mines conceptual domains from the Master Metaphor List, OpenCyc, and WordNet to identify metaphorical word pairs and their underlying domains~\cite{tian2024bridging}. For sarcasm, KSDGCN combines COMET-generated commonsense with sentiment and dependency structures to identify implausible relations~\cite{wang2025elevating}, while SD-APRR incorporates event consequences and human reactions through COMET's xEffect and xReact relations~\cite{min-etal-2023-just}. Similar strategies support implicit-hate understanding: KID-VLM constructs joint graphs from meme text, generated captions, and ConceptNet entities~\cite{garg-etal-2025-just}, and Wei et al.\ introduce codetype taxonomies through prompting or frozen-LLM representations~\cite{wei-etal-2025-cracking}. Although structured resources improve interpretability by making relevant relations explicit, they may be incomplete, noisy, or mismatched with the context in which high-level meanings arise.

\textbf{Parametric knowledge fusion} incorporates external knowledge during training, allowing models to internalize domain corpora, expert resources, teacher outputs, conceptual labels, or task-specific instruction data. AdMul derives basic-sense knowledge from word-sense disambiguation data and WordNet, then transfers it to metaphor detection through adversarial multitask learning, enabling MIP-style comparison between contextual and basic meanings without inference-time retrieval~\cite{zhang-liu-2023-adversarial}. Psyche-R1 instead constructs large-scale Chinese psychological training data from educational, clinical, platform-based, and LLM-distilled sources, and combines SFT with GRPO to improve empathy, psychological expertise, and reasoning~\cite{dai-etal-2026-psyche}. Compared with retrieval-based methods, parametric fusion enables more efficient inference and stable task specialization, but provides less transparent access to knowledge provenance and makes subsequent knowledge updates more difficult.

Overall, these three families differ primarily in where external knowledge is stored and when it is introduced. Retrieval-augmented methods access knowledge dynamically, symbolic methods explicitly represent structured relations, and parametric methods internalize knowledge during training. Their complementary strengths suggest that reliable HLS understanding may benefit from combining dynamic retrieval, interpretable symbolic structure, and task-specialized parametric knowledge.

\subsubsection{Context Modeling}

HLS understanding often depends on context beyond the current utterance, including discourse history, narrative progression, interaction trajectories, speaker identity, audience perspective, and participants' social or psychological states. Context modeling therefore seeks to explain how high-level meanings emerge, evolve, and become interpretable within broader communicative situations. 
Existing methods mainly capture two complementary forms of context: the temporal development of discourse and the participant-specific or broader social perspectives through which it is interpreted. 

The first direction performs \textbf{discourse-history modeling}, treating high-level meaning as something that unfolds across previous turns, narrative buildup, or interaction trajectories rather than being recoverable from an isolated utterance. Dialogue windows, chronological memory, sequential labeling, and state tracking are commonly used to capture emotional evolution, humor timing, sarcastic contrast, and persuasive adaptation. SENSE-7 models perceived empathy across sustained human--AI conversations by combining turn-level annotations with conversational history, user characteristics, and task information~\cite{suh2026sense}, showing that empathy judgments depend on conversational continuity and user expectations. Humor understanding similarly benefits from broader sequential context: StandUp4AI formulates stand-up humor detection as word-level sequence labeling to capture local discourse buildup~\cite{barriere-etal-2025-standup4ai}, while full-clip modeling outperforms short-segment modeling in football press conferences~\cite{10707160}. For sarcasm and sentiment analysis, Zhang et al.\ jointly model target and contextual utterances, speaker information, and interaction patterns, capturing contrasts between the current expression and the preceding discourse~\cite{zhang2023learning}. Collectively, these studies show that discourse history provides the temporal structure needed to interpret meanings that depend on accumulation, contrast, or change.

The second direction performs \textbf{participant-perspective and social-context modeling}, explaining high-level meaning by identifying who is speaking, whose perspective is being inferred, what social or psychological state each participant occupies, and what broader social setting shapes the interaction. 
In mental-health support conversations, Yang et al.\ model empathy as alignment between the Target's and Observer's appraisals, predicting whether the Observer's response reflects the Target's perspective~\cite{yang-jurgens-2024-modeling}. PPEP further treats empathy as perspective-dependent by predicting how one person responds to another's opposing-view story based on that person's own experience~\cite{chen-etal-2025-empathy}. These studies suggest that empathy cannot be determined from the target text alone, because its interpretation depends on the experiences and viewpoints of both participants. Participant-specific context is also important for humor: CIHR combines general humor patterns with static and dynamic speaker profiles derived from speaker attributes and historical posts, allowing the model to account for individualized expression styles~\cite{zhu2025cihr}. 
Extending perspective modeling to broader social settings, SocialStoryFrames infers perceived author intent and reader reception from both conversational context and community norms, covering causal explanations, predictions, character appraisals, value judgments, and affective responses to social-media narratives~\cite{mire-etal-2026-social}.

Overall, discourse-history modeling captures how HLS develops over time, whereas participant-perspective and social-context modeling captures how it varies across speakers, audiences, and communicative communities. 
Together, they show that reliable HLS understanding requires situating an expression within both the evolving interaction and the social perspectives through which it is interpreted.

\subsubsection{Multimodal Semantic Clue Integration}

HLS is frequently conveyed through combinations of textual, visual, acoustic, and temporal information. Because the intended meaning may not be recoverable from any single modality, multimodal HLS understanding requires models to identify and integrate complementary, conflicting, and implicit semantic clues. Existing methods mainly follow two directions: adaptive cross-modal fusion and semantic-bridge construction.

The first direction \textbf{adaptively fuses cross-modal semantic clues} through attention, graph structures, contrastive learning, gating, expert routing, or uncertainty-aware weighting. Rather than treating all modalities as equally informative, these methods distinguish their complementary and conflicting contributions. MMOE separates modal redundancy, uniqueness, and synergy through unimodal and multimodal predictions and assigns specialized experts to different interaction types~\cite{yu-etal-2024-mmoe}. For sarcasm detection, AMIF models textual, visual, and cross-modal incongruity and uses ambiguity guidance to select among conflicting clues~\cite{li2025amif}, while HCT-DMG dynamically identifies a primary modality and hierarchically integrates auxiliary information to suppress misleading signals~\cite{wang2023cross}. MGMCF captures both global and local interactions among text, images, and visual objects, allowing subtle object-level evidence to contribute to meme and metaphor understanding~\cite{zheng2025multi}. Adaptive integration also extends to other HLS tasks: APCL constructs persuasion and anti-persuasion prompt pairs for contrastive multimodal propaganda detection~\cite{cui2023multimodal}, and multimodal empathy detection combines lexical information with complementary acoustic-prosodic cues~\cite{chen2024detecting}. These studies show that effective multimodal integration requires not only combining features, but also determining which clues are reliable, complementary, or semantically decisive in a given context.

The second direction \textbf{constructs semantic bridges for implicit multimodal clues}, translating heterogeneous evidence into captions, rationales, reasoning steps, or structured representations before final prediction. Some methods guide models through decomposed reasoning: C4MMD performs staged CoT reasoning over image, text, and cross-modal meaning~\cite{xu2024c4mmd}; PunchBench uses simple-to-complex questions to integrate visual and textual clues before evaluating punchlines~\cite{ouyang2025punchbench}; and Commander-GPT routes inputs to specialized analyses of context, sentiment, rhetoric, facial expressions, images, and scene text before aggregating their outputs~\cite{zhang2025commandergpt}. Other methods construct task-specific semantic structures. ImaRA represents multimodal metaphors through imaginative frames involving source and target domains, domain inconsistency, and attribute similarity~\cite{tian-etal-2025-imara}, while D-HUMOR explains meme humor through its joke, narrative, emotional effect, dark attributes, and target~\cite{kasu2025d}. MiDRE balances direct image--text evidence against LVLM-generated rationales through a gating mechanism~\cite{jana2025midre}, and M3H combines OCR text, commonsense knowledge, and generated reasoning about figurative meaning, causality, and mental states~\cite{mazhar2025figurative}. Semantic bridging can also textualize non-textual information: Pro-Cap generates targeted captions for socially relevant visual clues~\cite{cao2023procap}, whereas SpeechCueLLM converts vocal cues into natural-language descriptions~\cite{wu-etal-2025-beyond}. Although these bridges make implicit evidence more accessible and interpretable, generated captions and rationales may introduce hallucinated or biased information, making reliability control essential.

Overall, adaptive fusion integrates multimodal clues within learned representations, whereas semantic-bridge construction first converts them into explicit linguistic or structured forms. The two directions are complementary: the former emphasizes flexible weighting and interaction among modalities, while the latter improves the accessibility and interpretability of otherwise implicit evidence.

\subsubsection{Data-Efficient Learning}

A persistent challenge in HLS understanding is that high-quality task data are often scarce and expensive to annotate. HLS labels may require pragmatic inference, affective interpretation, cultural knowledge, or theory-informed judgment. Existing methods address this limitation through three main strategies: enriching each labeled instance with denser supervision, introducing semantic priors to reduce annotation dependence, and adapting pretrained models through lightweight transfer.

The first strategy \textbf{enhances supervision through auxiliary structures}, allowing each label to provide richer learning signals. In empathy detection, CLSN jointly predicts empathy and communicative intent, using intent labels to characterize why an utterance is empathetic~\cite{jiang2023empathyintent}. Metaphor methods similarly exploit related tasks: basic-sense discrimination transfers word-sense knowledge to metaphor detection~\cite{zhang-liu-2023-adversarial}, while joint learning across related figurative phenomena shares their linguistic structure~\cite{badathala-etal-2023-match}. Curriculum learning further improves data utilization by gradually introducing more difficult metaphor examples~\cite{jia-li-2024-metaphor}. For sarcasm, auxiliary sentiment and emotion objectives can incorporate unlabeled data through semi-supervised learning~\cite{irfan2025enhancing}, while multitask architectures separate task-shared and task-specific information between sarcasm and sentiment analysis~\cite{zhang2023learning}. Contrastive methods provide another source of dense supervision: APCL constructs persuasion and anti-persuasion prompt pairs with hard negatives~\cite{cui2023multimodal}, whereas CLCL contrasts figurative and literal uses and organizes them through a dynamic curriculum~\cite{zhou2023clcl}. These methods make limited labels more informative by embedding them within related tasks, semantic contrasts, or difficulty-aware learning schedules.

The second strategy \textbf{injects HLS priors into reasoning}, reducing reliance on large annotated datasets through linguistic theories, psychological models, commonsense knowledge, or explicit reasoning structures. In metaphor understanding, MisNet operationalizes MIP and SPV by comparing contextual and dictionary-derived basic meanings~\cite{zhang2022misnet}; related work converts SPV, MIP, and Conceptual Metaphor Theory into knowledge graphs and scaffolding questions for zero-shot reasoning~\cite{tian-etal-2024-theory}. ImaRA extends theory-guided reasoning to low-resource multimodal metaphors by retrieving imaginative frames that encode source--target mappings and attribute similarity~\cite{tian-etal-2025-imara}. For empathy, appraisal theory provides structure for modeling perspective alignment between targets and observers~\cite{yang-jurgens-2024-modeling}. Sarcasm methods instead decompose interpretation into explicit entities, discrepancies, pragmatic cues, or specialized subtasks~\cite{jiang2026satiredecoder,yao2025sarcasm,lee-etal-2025-pragmatic,zhang2025commandergpt}. Similar pragmatic scaffolds have been applied to implicit toxic-language detection through inference steps grounded in Relevance Theory~\cite{chen-wang-2025-pragmatic}. Such priors are especially useful in zero-shot and few-shot settings, although their reliability depends on whether the selected theories, retrieved knowledge, and generated reasoning paths fit the input.

The third strategy uses \textbf{lightweight transfer and parameter-efficient adaptation}. Rather than fully fine-tuning large models on limited HLS data, these methods freeze most pretrained parameters and optimize small task-specific components or transfer representations learned from broader data. ImageMet combines task-specific CoT prompting with LoRA fine-tuning for visual metaphor captioning and VQA~\cite{kundu2025looking}. AdS-CLIP inserts adapters into the upper layers of frozen CLIP encoders and shares adapter states across text and vision to support sarcasm-specific cross-modal learning~\cite{jana2026adapter}. For spoken HLS tasks, task-agnostic contrastive pretraining can align speech and text representations before limited downstream fine-tuning~\cite{zufle-niehues-2025-contrastive}. These approaches reduce computational cost and overfitting, but remain dependent on whether the pretrained backbone already contains the required cultural, pragmatic, affective, or multimodal knowledge.

Overall, the three strategies improve data efficiency from complementary perspectives: supervision enhancement extracts more information from available labels, semantic priors constrain learning and reasoning, and lightweight transfer reuses knowledge stored in pretrained models.

\subsection{Evaluation}
\label{sec:hlsu_eval}

Evaluating HLS understanding requires more than measuring whether a model predicts the correct label. Depending on the task, evaluation may also need to examine semantic mechanisms, explanation quality, multimodal grounding, annotator subjectivity, and generalization. We organize existing approaches into five complementary categories.

\subsubsection{Gold-Label-Based Evaluation}

The most common approach compares model predictions with human-, expert-, or dataset-annotated labels. HLS understanding is formulated as classification, detection, ranking, or regression and evaluated using accuracy, F1, AUC, correlation, RMSE, or related metrics. For example, empathy in speech can be evaluated as empathetic-versus-neutral classification using accuracy and F1, supplemented with statistical analysis of acoustic-prosodic features~\cite{chen2024detecting}. MHSDB standardizes feature extraction, data splits, and accuracy/F1 reporting across English and Hindi multimodal humor and sarcasm datasets, improving comparability across methods~\cite{dong2025mhsdb}. SafePersuasion similarly evaluates the distinction between rational persuasion and manipulation through binary and multi-label detection~\cite{kong-etal-2025-safepersuasion}. Gold-label evaluation is scalable and reproducible, but it often compresses subjective, culturally dependent, or multidimensional meanings into a single target label.

\subsubsection{Fine-Grained Evaluation}

Fine-grained evaluation decomposes HLS understanding into interpretable semantic components, such as communicative intent, span localization, source--target mapping, explanation adequacy, and mental-state tracking. EmpEval evaluates empathy through 16 expressed intents and four perceived dimensions, comparing model predictions with human judgments using standard classification metrics~\cite{xu2024empeval}. DocMSU evaluates both document-level sarcasm detection and the localization of sarcastic textual spans and visual objects through specialized overlap and error metrics~\cite{Du_Nan_Zhang_Xie_Xu_Fan_Cui_Tao_Jiang_2024}. For metaphor, Metaphor Mapping Identification tests whether models recover source and target domains from image--text advertisements using exact match, BERTScore, and human evaluation~\cite{zhang2025towards}. Chumor evaluates whether explanations adequately account for Chinese jokes through classification metrics, human baselines, preference tests, and error analysis~\cite{he-etal-2025-chumor}. Persuasive and negotiation dialogue benchmarks further assess desire, belief, intention, strategy, and consistency across turns~\cite{yu2025persuasivetom,chan2024negotiationtom}. These evaluations reveal which semantic components a model understands, although they require richer annotations and must account for partially correct or ambiguous structured outputs.

\subsubsection{Judge-Based Evaluation}

When HLS outputs are open-ended, judge-based evaluation relies on humans, experts, LLMs, or simulated agents to assess explanations, emotional reasoning, cultural interpretation, and social cognition. EmotionQueen combines GPT-4 scoring with human validation to evaluate the recognition of key events, mixed events, implicit emotions, and intentions~\cite{chen-etal-2024-emotionqueen}. SAGE uses simulated agents with evolving emotional states and evaluates supportive-dialogue understanding through emotion trajectories, empathy correlations, model--human consistency, and cross-agent robustness~\cite{zhang-etal-2026-sentient}. For humor, human ratings of explanation accuracy and completeness have been used to compare LLM judges with automatic metrics, revealing differences across joke topics and pun types that binary detection may conceal~\cite{loakman-etal-2025-comparing}. V-FLUTE combines entailment labels with explanation-aware metrics and expert error analysis for figurative visual reasoning~\cite{saakyan-etal-2025-understanding}, while FLUB integrates answer accuracy, fallacy-type classification, GPT-4 scoring, and human verification for Chinese fallacy understanding~\cite{li2024llms}. Judge-based methods extend evaluation from answer correctness to reasoning adequacy and semantic coherence, but their conclusions depend on the reliability and cultural alignment of the judges themselves.

\subsubsection{Perception--Reasoning-Decomposed Multimodal Evaluation}

Multimodal evaluation often separates lower-level perception from higher-level semantic reasoning to determine whether models genuinely integrate modalities or rely on textual shortcuts. EchoMind decomposes empathetic speech understanding into spoken-content recognition, vocal-cue perception, and integrated reasoning, showing that speech-language models perform better on linguistic content than on paralinguistic reasoning~\cite{zhou2025echomindinterrelatedmultilevelbenchmark}. MEDIC evaluates counseling empathy using textual, acoustic, and visual information through accuracy and macro-F1~\cite{zhu2023medic}. For visual humor, PunchBench separates perception and punchline reasoning across multiple QA formats and introduces synonymous and antonymous captions to reduce shortcut reliance~\cite{ouyang2025punchbench}. PixelHumor evaluates comic understanding through humor detection, style classification, open-ended interpretation, and panel or text ordering, combining automatic metrics with human ratings~\cite{ryan-etal-2025-humor}. SarcNet provides separate image, text, and multimodal sarcasm labels, enabling modality-specific evaluation~\cite{yue-etal-2024-sarcnet}. ImageMet similarly evaluates visual metaphors through captioning and VQA with both automatic and human metrics; their weak correlation illustrates the difficulty of measuring metaphorical understanding automatically~\cite{kundu2025looking}. By decomposing perception and reasoning, these benchmarks help identify whether errors originate from recognizing multimodal clues or integrating them into high-level meanings.

\subsubsection{Evaluation of Subjectivity, Robustness, and Generalization}

Because HLS judgments vary across people, cultures, prompts, and domains, evaluation should also examine disagreement, shortcut reliance, and distribution shift. For subjectivity, SENSE-7 uses turn-level ratings along seven empathy dimensions and shows that perceived empathy varies with conversational context and user characteristics~\cite{suh2026sense}; MultiPICo preserves disaggregated irony annotations and annotator metadata to evaluate alignment with different demographic groups~\cite{casola-etal-2024-multipico}. For robustness, literal-paraphrase adversarial tests and prompt variations expose metaphor models' sensitivity to lexical overlap and sentence length~\cite{sanchez-bayona-agerri-2025-metaphor}, while minimally contrastive image pairs in HumorDB test whether models attend to humor-relevant visual differences~\cite{jain2025humordb}. For generalization, cross-dataset sarcasm evaluation reveals dependence on dataset-specific styles and differences between author labels and third-party judgments~\cite{jang-frassinelli-2024-generalizable}, while metaphor studies examine cross-lingual and cross-dataset transfer~\cite{aghazadeh-etal-2022-metaphors}. These perspectives treat disagreement and distributional variation as central properties of HLS evaluation rather than incidental annotation noise.

%% file: 04_2_high-level_semantic_intelligence_gen.tex
\section{High-Level Semantics Generation}
\label{sec:hlsg}

HLS generation aims to achieve the intended semantic effect by coherently integrating semantic clues within a given context. 
Therefore, its central challenge lies not merely in generating fluent sentences, realistic entities, or acoustic patterns, but in coordinating these elements into a coherent whole that evokes the intended high-level meaning.

\begin{figure*}[!t]
  \centering

  \definecolor{HLSRootFill}{RGB}{255,226,174}
  \definecolor{HLSRootDraw}{RGB}{255,183,77}

  \definecolor{DataHeadFill}{RGB}{255,194,199}
  \definecolor{DataHeadDraw}{RGB}{255,140,150}
  \definecolor{DataLeafFill}{RGB}{253,233,235}
  \definecolor{DataLeafDraw}{RGB}{248,205,210}

  \definecolor{ApproachHeadFill}{RGB}{178,204,248}
  \definecolor{ApproachHeadDraw}{RGB}{112,163,246}
  \definecolor{ApproachLeafFill}{RGB}{226,236,252}
  \definecolor{ApproachLeafDraw}{RGB}{203,220,249}

  \definecolor{EvalHeadFill}{RGB}{198,184,248}
  \definecolor{EvalHeadDraw}{RGB}{145,124,226}
  \definecolor{EvalLeafFill}{RGB}{235,231,252}
  \definecolor{EvalLeafDraw}{RGB}{217,208,247}

  \tikzset{
    root node/.style={
      draw=HLSRootDraw,
      fill=HLSRootFill,
      align=center,
      thin,
      rounded corners=3,
    },
    data head/.style={
      draw=DataHeadDraw,
      fill=DataHeadFill,
      align=center,
      thin,
      text width=3cm,
      rounded corners=3,
    },
    data leaf/.style={
      draw=DataLeafDraw,
      fill=DataLeafFill,
      align=center,
      thin,
      text width=8.8cm,
      rounded corners=3,
    },
    approach head/.style={
      draw=ApproachHeadDraw,
      fill=ApproachHeadFill,
      align=center,
      thin,
      text width=3cm,
      rounded corners=3,
    },
    approach leaf/.style={
      draw=ApproachLeafDraw,
      fill=ApproachLeafFill,
      align=center,
      thin,
      text width=8.8cm,
      rounded corners=3,
    },
    evaluation head/.style={
      draw=EvalHeadDraw,
      fill=EvalHeadFill,
      align=center,
      thin,
      text width=3cm,
      rounded corners=3,
    },
    evaluation leaf/.style={
      draw=EvalLeafDraw,
      fill=EvalLeafFill,
      align=center,
      thin,
      text width=8.8cm,
      rounded corners=3,
    }
  }

  \begin{forest}
    nonleaf/.style={
      font=\bfseries\scriptsize
    },
    for tree={
      font=\scriptsize,
      text=black,
      anchor=west,
      inner sep=2pt,
      l sep=7.5pt,
      s sep=4pt,
      fit=tight,
      grow'=east,
      edge={
        ultra thin,
        draw=black!65
      },
      parent anchor=east,
      child anchor=west,
      if n children=0{}{
        nonleaf
      },
      edge path={
        \noexpand\path[\forestoption{edge}]
        (!u.parent anchor) -- +(5pt,0) |- (.child anchor)
        \forestoption{edge label};
      },
      if={isodd(n_children())}{
        for children={
          if={equal(n,(n_children("!u")+1)/2)}{
            calign with current
          }{}
        }
      }{}
    }
    [
      {\rotatebox{90}{HLS Generation (Section~\ref{sec:hlsg})}},
      root node,
      text width=0.5cm,
      font=\bfseries\scriptsize,
      inner sep=2pt,
      minimum height=1.8cm,
      minimum width=0.45cm
      [
        Data \\(Section~\ref{sec:hlsg_data}),
        data head,
        for descendants={data leaf}
        [
          Theory-Guided Semantic Structuring
        ]
        [
          LLM-Assisted Data Synthesis
        ]
        [
          Contextual and Multimodal Grounding
        ]
        [
          Quality Control
        ]
      ]
      [
        Approaches \\(Section~\ref{sec:hlsg_approach}),
        approach head,
        for descendants={approach leaf}
        [
          Task-Specific and Multi-Objective Training
        ]
        [
          Feedback-Guided Optimization
        ]
        [
          Mechanism-Aligned Semantic Modeling
        ]
        [
          Synergistic Interactions among High-Level Semantics
        ]
        [
          Creative and Diverse Generation
        ]
        [
          Multimodal Semantic Integration and Generation
        ]
        [
          Strategy-Guided Controllable Generation
        ]
        [
          External Knowledge Integration
        ]
        [
          Personalized and User-Adaptive Generation
        ]
        [
          Data-Efficient Learning
        ]
      ]
      [
        Evaluation \\(Section~\ref{sec:hlsg_eval}),
        evaluation head,
        for descendants={evaluation leaf}
        [
          Reference-Based and Semantic Similarity Evaluation
        ]
        [
          Judgment-Based Evaluation
        ]
        [
          Semantic Attribute and Control Evaluation
        ]
        [
          Multimodal-Consistency Evaluation
        ]
        [
          Diversity and Creativity Evaluation
        ]
        [
          Extrinsic and Downstream Evaluation
        ]
      ]
    ]
  \end{forest}

  \caption{
    Taxonomy of High-Level Semantic (HLS) generation methods, covering
    data construction, generation approaches, and evaluation protocols.
  }
  \label{fig:generation-taxonomy}
\end{figure*}
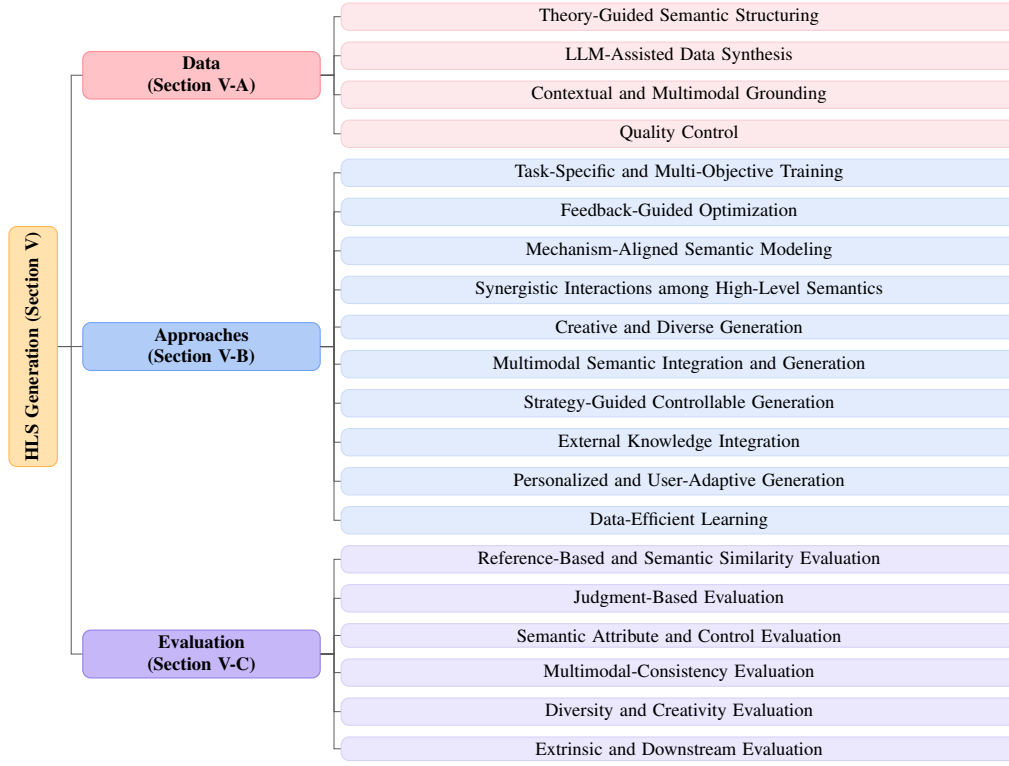

\subsection{Data}
\label{sec:hlsg_data}

Data for HLS generation must specify not only the content to be generated, but also the intended semantic effect, its underlying mechanism, and the context in which it should be realized. Existing work improves generation data along four complementary dimensions: structuring latent semantic mechanisms, expanding coverage through LLM-assisted synthesis, grounding generation in contextual and multimodal situations, and strengthening validity through quality control. 

\subsubsection{Theory-Guided Semantic Structuring}

HLS generation resources increasingly represent latent semantic mechanisms as control variables or intermediate supervision. These structures encode incongruity, source--target mappings, sentiment contrasts, emotion causes, communicative intents, or interaction strategies, enabling models to control how a high-level meaning is realized. For humor, pun-generation resources annotate pun words, alternative words, word senses, ambiguity, and surprise~\cite{mittal-etal-2022-ambipun,sun2022context,tian-etal-2022-unified}, while IRCoT and HUMORCHAIN decompose visual humor into image description, incongruity detection, resolution, plausibility assessment, and humor strategy~\cite{tanaka-etal-2024-content,zhang2025humorchain}. Metaphor resources encode tenor--vehicle or source--target mappings, grounding, intended meaning, paraphrases, and selectional-preference violations to preserve mapping fidelity during generation~\cite{zhang-etal-2024-gome,koushik2025mindseyemultifacetedreward,mao-etal-2024-metapro,chen-etal-2024-merely}. Sarcasm resources instead capture targets, sentiment inconsistencies, multimodal mismatches, speaker relations, and explanatory rationales~\cite{goel-etal-2025-target,tmm25sarcasmexp}. For interpersonal generation, semantic structures are often derived from psychological and social-science theories. Empathy resources represent cognitive and affective empathy, emotion causes, communicative intents, appraisal variables, support strategies, cognitive reframing, and counseling principles~\cite{zhou-etal-2023-case,zhao-etal-2023-dont,wang-etal-2022-empathetic,chen-etal-2022-emphi,10.1145/3627673.3679687,xiao-etal-2024-healme,yao2025empathy,bn2025pursuit}, whereas persuasion resources annotate strategies, politeness, resistance, conflict styles, negotiation phases, mental states, and speaker stances~\cite{samad-etal-2022-empathetic,donate2savealife,jin-etal-2024-persuading,chi24audiencedepolarisation,zhang2026persuasion,priya-etal-2024-trip}. Such structures provide explicit semantic plans for controllable HLS generation rather than supervising surface outputs alone.

\subsubsection{LLM-Assisted Data Synthesis}

When naturally occurring examples are scarce, HLS datasets can be expanded through LLM-based generation and simulation. Unlike ordinary augmentation that primarily paraphrases existing instances, these methods attempt to preserve or manipulate latent pragmatic functions such as humor, metaphorical mapping, sarcastic contrast, empathetic support, and persuasive strategy. Existing pipelines mainly employ controlled rewriting, theory-guided decomposition, and multi-agent simulation.

The first approach uses \textbf{LLM-based rewriting or controlled generation}, in which existing utterances, scenarios, or dialogue contexts serve as seeds for generating outputs under specified semantic constraints. Feedback-driven humor distillation, for example, converts literal sentences into meaning-preserving humorous paraphrases, producing paired data for student-model training~\cite{ravi-etal-2024-small}. Synth-Empathy and SYNTHEMPATHY similarly expand empathy data through response generation, scenario brainstorming, therapy-style rewriting, and explanation--response construction~\cite{liang2024synthempathy,chen2025synthempathy}. This approach is scalable when a target phenomenon can be adequately specified through prompts and demonstrations, although its quality depends on whether the generated outputs preserve the intended semantic effect.

A second approach performs \textbf{theory-guided synthetic generation}, decomposing the target phenomenon into intermediate semantic variables before generating the final output. For visual metaphor construction, LLMs identify source and target domains, implicit meanings, candidate objects, and visual elaborations before invoking image-generation models~\cite{chakrabarty-etal-2023-spy}. GOME and The Mind's Eye further combine chain-of-thought prompting, rhetorical knowledge, and iterative reward feedback to refine source--target--meaning structures and visual prompts~\cite{zhang-etal-2024-gome,koushik2025mindseyemultifacetedreward}. Humor synthesis follows a similar logic: IRCoT decomposes generation into image description, incongruity identification, resolution generation, and caption selection~\cite{tanaka-etal-2024-content}, while HUMORCHAIN incorporates entity recognition, plausibility assessment, and theory-linked humor strategies~\cite{zhang2025humorchain}. Explicit decomposition improves controllability by grounding synthetic samples in identifiable semantic mechanisms.

A third approach uses \textbf{multi-agent simulation} for interactional phenomena whose semantic effects emerge over multiple turns. Agents are assigned social roles, communicative goals, user profiles, or monitoring functions and jointly simulate dialogue trajectories. Empathy research uses multi-agent settings to construct sticker-based conversations and clinically grounded support interactions with diverse emotional developments~\cite{zhang-etal-2024-stickerconv,bn2025pursuit}. Persuasion datasets employ this paradigm more extensively: DailyPersuasion generates scenarios, strategies, and conversations with GPT-4~\cite{jin-etal-2024-persuading}, while multi-LLM frameworks assign persuader, persuadee, monitor, annotator, and regulator roles to simulate belief change, resistance, and strategy adaptation~\cite{ma-etal-2025-communication}. PersuaSim further reinforces speaker profiles and multilingual prompts across generation rounds to produce culturally grounded dialogues~\cite{ma-etal-2025-enhancing}. Although simulation broadens interaction coverage, synthetic data remain vulnerable to semantic drift, templated behavior, hallucinated rationales, and insufficient sensitivity to social or clinical risks.

\subsubsection{Contextual and Multimodal Grounding}

\textbf{Contextual grounding} situates HLS generation within dialogue history, speaker identity, user profiles, cultural background, mental states, stances, or communicative goals. 
In humor, context-situated pun and image-conditioned caption datasets require jokes to fit a particular textual or visual setting~\cite{sun2022context,tanaka-etal-2024-content,11209232,zhang2025humorchain}. 
Empathy resources use persona retrieval, personality traits, MBTI signals, user histories, PTSD personas, and perspective-taking stages to generate responses aligned with individual emotional needs~\cite{huang2024generating,wu-etal-2025-traits,cai2024pecer,bn2025pursuit,10.1145/3774904.3793022}. 
Persuasion resources similarly condition generation on culture, stance, hidden beliefs, preferences, and negotiation phases~\cite{ma-etal-2025-enhancing,zhang2026persuasion,moore2025large,priya-etal-2024-trip,Priya_Chigrupaatii_Firdaus_Ekbal_2025}. 
These datasets shift generation from producing generally plausible outputs toward selecting semantic strategies appropriate to particular participants and situations.

\textbf{Multimodal grounding} further incorporates images, videos, speech, stickers, avatars, memes, and audiovisual clues. Visual humor and metaphor resources cover video explanations, meme captions, meme generation, and visual metaphor construction, where meaning emerges from interactions among text, image, sound, and composition~\cite{ko-etal-2023-language,hyun-etal-2024-smile,hwang-shwartz-2023-memecap,mm24xmecap,www24memecraft,chakrabarty-etal-2023-spy,koushik2025mindseyemultifacetedreward}. Several pipelines first generate textual explanations, incongruity descriptions, or visual prompts and then pair them with image or video generation models~\cite{chakrabarty-etal-2023-spy,zhang-etal-2024-gome,tanaka-etal-2024-content}. Sarcasm data include image--text posts, audiovisual dialogues, and sarcastic speech because irony may be signaled by cross-modal mismatch or prosody~\cite{jing-etal-2023-multi,goel-etal-2025-target,tmm25sarcasmexp,li25b_ssw}. Empathy and persuasion resources also extend to talking-head generation, sticker responses, multi-party support, social-cause memes, mediation, negotiation, and bargaining~\cite{11129652,10.1145/3746027.3762031,lin2025e3rg,wang2025empathyomnienablingempathetic,hu-etal-2025-chain,liu2024generative,zhang-etal-2024-stickerconv,zhu-etal-2022-multi,chi24audiencedepolarisation,shea-yu-2024-fairness,Zahedi_Sengupta_Kambhampati_2024}. These resources broaden HLS generation from sentence-level outputs to situated communicative artifacts.

\subsubsection{Quality Control}

Because HLS data are subjective, ambiguous, and socially sensitive, quality control must assess pragmatic validity rather than surface fluency alone. Existing pipelines generally combine phenomenon-specific human validation with automatic filtering and iterative refinement to ensure that generated examples are semantically effective, contextually appropriate, and safe.

\textbf{Human and expert validation} applies criteria tailored to different HLS phenomena. Humor data are evaluated for funniness, creativity, relevance, coherence, informativeness, and safety~\cite{ko-etal-2023-language,a78fdd6694d54f5491f7393dfc7529e1,mm24xmecap,mittal-etal-2022-ambipun,sun2022context,zhang-wan-2022-mover}. Metaphor resources verify whether generated outputs preserve the intended source--target mapping and meaning rather than merely exhibiting visual appeal or linguistic fluency~\cite{chakrabarty-etal-2023-spy,zhang-etal-2024-gome,koushik2025mindseyemultifacetedreward,mao-etal-2024-metapro}. For socially sensitive tasks, validation additionally considers potential harm. Sarcasm data are checked for targets, explanation quality, perceived sarcasm, and prosodic naturalness~\cite{goel-etal-2025-target,kumar-etal-2022-become,li25b_ssw,veselovsky2023generating}; empathy resources involve psychologists, experts, or trained annotators to assess emotional appropriateness, therapeutic plausibility, identity consistency, and safe support~\cite{bn2025pursuit,xiao-etal-2024-healme,shi-etal-2025-beyond,10.1145/3774904.3793022}; and persuasion data are validated for strategic plausibility, politeness, fairness, stance consistency, and argumentative support~\cite{samad-etal-2022-empathetic,donate2savealife,ma-etal-2025-communication,ma-etal-2025-enhancing,priya-etal-2024-trip,chi24audiencedepolarisation}.

\textbf{Automatic filtering and iterative refinement} embed quality control into a generate--evaluate--revise pipeline. These methods generate multiple candidates and use discriminators, reward or preference models, human feedback, or model-based criteria to select or revise them. Humor pipelines apply such mechanisms to improve funniness, relevance, and rhetorical appropriateness~\cite{ko-etal-2023-language,11209232,zhang2025humorchain}, while empathy synthesis filters generic, unsafe, or formulaic responses and promotes greater diversity~\cite{liang2024synthempathy,chen2025synthempathy}. Metaphor generation further uses iterative self-evaluation to verify the source, target, and intended meaning before accepting a visual prompt~\cite{koushik2025mindseyemultifacetedreward}. Together, human validation and iterative filtering help ensure that scalable data construction preserves the intended HLS effects.

\subsection{Approaches}
\label{sec:hlsg_approach}

\subsubsection{Task-Specific and Multi-Objective Training}

A central challenge in HLS generation is that standard likelihood-based objectives primarily favor fluency and local plausibility, whereas HLS tasks require outputs to satisfy global communicative and semantic constraints. 
Depending on the task, these constraints may involve funniness, metaphorical coherence, sarcasm recognizability, emotional validation, persuasiveness, credibility, safety, or fairness. Existing work therefore extends likelihood training with task-specific auxiliary objectives and multi-objective formulations.

One direction introduces \textbf{task-specific auxiliary objectives} that explicitly represent semantic or pragmatic properties of the target phenomenon. For example, PGCL separates pun-structure satisfaction from general generation quality, allowing structural constraints to guide whether the generated output constitutes a valid pun~\cite{chen-etal-2024-u}. In empathetic response generation, CASE jointly models response generation, emotion prediction, diversity, and cognitive--affective alignment~\cite{zhou-etal-2023-case}, while SEEK combines generation with emotion--intent tagging, response-level emotion and intent prediction, dialogue emotion recognition, and frequency-aware cross-entropy~\cite{wang-etal-2022-empathetic}. These auxiliary objectives expose task-relevant structure that token-level likelihood alone does not capture.

Another direction formulates HLS generation as \textbf{multi-objective optimization}. In persuasion and negotiation, models may jointly optimize empathy, politeness, persuasiveness, strategy consistency, persona alignment, coherence, diversity, and fairness-aware outcomes~\cite{mishra-etal-2023-pal,samad-etal-2022-empathetic,donate2savealife,priya-etal-2024-trip,Priya_Chigrupaatii_Firdaus_Ekbal_2025,shea-yu-2024-fairness}. Such formulations better reflect the multidimensional nature of HLS quality, but they also introduce trade-offs: improving persuasiveness may reduce fairness, while increasing stylistic intensity may weaken semantic fidelity or safety.

Overall, task-specific and multi-objective training extends likelihood-based learning with explicit semantic, affective, strategic, and social constraints. Its central challenge is to balance these objectives without allowing easily optimized surface properties to dominate the intended HLS effect.

\subsubsection{Feedback-Guided Optimization}

Feedback-guided optimization evaluates generated outputs and uses the resulting signals to update model parameters, select candidates, revise responses, or guide subsequent interactions. This closed-loop formulation is particularly important for HLS generation because many failures are semantic rather than grammatical: an output may be fluent yet insufficiently humorous, metaphorically incoherent, emotionally inappropriate, unpersuasive, unsafe, or inconsistent with the context. Existing methods incorporate feedback at three stages: training, inference, and interaction. 

The first direction uses \textbf{training-time feedback for parameter optimization}, converting human preferences, model judgments, task-specific rubrics, or discriminator scores into learning signals. 
In humor generation, DPO improves caption selection and diversity under certain distribution shifts~\cite{zhang2024humor}, while PGCL combines pun-structure constraints with triplet-based preference optimization~\cite{chen-etal-2024-u}. Other methods derive feedback from teacher critiques, ranking objectives, humor discriminators, or rationale-enhanced self-learning~\cite{ravi-etal-2024-small,11209232,wang2025innovative}. Empathetic generation typically requires multidimensional feedback: EmpRL defines rewards over emotional reaction, interpretation, and exploration~\cite{ma2025emprl}, while Kardia-R1 and FPEMF incorporate criteria such as relevance, emotional understanding, strategy effectiveness, persona consistency, and safety~\cite{10.1145/3774904.3793022,shi-etal-2025-beyond}. More recent GRPO-based methods optimize groups of sampled responses using multiple rubric-based rewards rather than relying only on fixed preference pairs. PERM, for example, constructs psychology-grounded rewards from supporter, seeker, and bystander perspectives before applying GRPO~\cite{wang2026permpsychologygroundedempatheticreward}. Feedback may also come from task-specific detectors, as in sarcasm-aware TTS, where a recognizer provides auxiliary losses for generating identifiable acoustic and semantic sarcasm cues~\cite{li25b_ssw}. These methods make subjective HLS qualities trainable, although their effectiveness depends on the reliability of the preferences, rubrics, and evaluators used.

The second direction applies \textbf{inference-time feedback for selection and refinement} without further parameter updates. HUMORCHAIN uses a humor discriminator to evaluate generated captions and trigger regeneration~\cite{zhang2025humorchain}. The Mind's Eye evaluates metaphorical images through source--target--meaning analysis, semantic similarity, and multimodal alignment, then incorporates the resulting critiques into iterative prompt revision~\cite{koushik2025mindseyemultifacetedreward}. PC-CRS alternates between a critic that checks candidate explanations against item information and a refiner that corrects inconsistencies~\cite{qin-etal-2024-beyond}. DMNA similarly uses multiple critics to assess dialogue quality and converts their judgments into actionable feedback for response regeneration~\cite{liu-etal-2025-dual}. These methods are useful when local fluency conceals deeper semantic failures, although self-refinement may reinforce errors when the generator and critic share the same incorrect assumptions.

The third direction uses \textbf{interaction-level feedback from agents, users, or simulated outcomes}. Instead of assigning a static score to an isolated output, these methods evaluate how generated content affects subsequent interpretations, user states, or dialogue trajectories. CoMet improves metaphor generation through self-play, using game outcomes and counterpart interpretations as feedback~\cite{xu-zhong-2025-comet}. Multi-agent persuasion pipelines assign monitoring, regulation, annotation, and post-processing roles to identify strategically weak or contextually inappropriate utterances~\cite{ma-etal-2025-communication}. ToMMA evaluates whether generated responses adequately address the persuadee's inferred beliefs and desires~\cite{zhang2026persuasion}. PersuGPT simulates future dialogue paths with a learned user model and converts comparisons of long-term persuasive outcomes into DPO preferences~\cite{jin-etal-2024-persuading}. ToM-Agent compares predicted and observed counterpart utterances and uses counterfactual reflection to revise its estimates of beliefs, desires, and intentions~\cite{yang2025large}. Such feedback is particularly relevant to empathy, persuasion, and negotiation, whose effectiveness often becomes evident only through later responses.

Overall, feedback-guided optimization shifts HLS generation from one-shot production toward closed-loop semantic alignment. Training-time feedback updates model parameters, inference-time feedback supports selection and revision, and interaction-level feedback captures longer-term communicative effects. Its effectiveness nevertheless depends on feedback reliability: human judgments are subjective and costly, LLM judges may exhibit cultural or positional biases, and discriminators may reward superficial proxies~\cite{ravi-etal-2024-small,zhang2025humorchain,ma-etal-2025-pun2pun}. Reliable feedback-guided generation therefore requires evaluators and revision procedures that closely match the intended semantic mechanism and communicative effect.

\subsubsection{Mechanism-Aligned Semantic Modeling}

HLS generation often requires reasoning over latent communicative variables that are not directly expressed in the input. 
Rather than mapping surface context directly to an output, mechanism-aligned methods introduce intermediate structures representing incongruity, ambiguity, conceptual mappings, emotional causes, communicative intentions, mental states, and anticipated reactions. 
These structures provide interpretable interfaces between input understanding and output generation. 
Existing methods can be broadly organized according to whether they model content-internal semantic mechanisms, affective and causal states, or interlocutor-level mental states.

The first direction performs \textbf{content-internal semantic mechanism structuring}, decomposing high-level effects into relational components or manipulable semantic variables to guide generation and subsequent refinement.
For visual humor, IRCoT separates image description, incongruity extraction, resolution generation, caption generation, and candidate selection~\cite{tanaka-etal-2024-content}, while HUMORCHAIN combines entity recognition, incongruity detection, plausibility assessment, and theory-derived humor strategies~\cite{zhang2025humorchain}. 
Pun generation similarly represents ambiguity through sense-specific contexts, contrasting meanings, support words, and pragmatic effects. AMBIPUN retrieves contextual words for both senses of a pun~\cite{mittal-etal-2022-ambipun}; a unified framework operationalizes ambiguity, distinctiveness, and surprise through context selection and word-level prediction~\cite{tian-etal-2022-unified}; and the CVO framework reconstructs cross-lingual puns through meaning constants, contextual triggers, support words, and pragmatic effects~\cite{ma-etal-2025-pun2pun}. Metaphor-oriented systems instead employ source--target--meaning or tenor--vehicle--grounding representations to preserve conceptual mappings during generation~\cite{koushik2025mindseyemultifacetedreward,zhang-etal-2024-gome}. LLM--diffusion pipelines further decompose linguistic metaphors into implicit meanings and visual objects before image synthesis~\cite{chakrabarty-etal-2023-spy}, while visual metaphor systems organize reasoning through scene construction, visual observation, abstract interpretation, and refinement~\cite{kundu2025looking}. 
For narrative generation, AGENTS' ROOM draws on narrative theory to assign conflict, character, setting, plot, and successive stages of the narrative arc to specialized agents coordinated through a shared scratchpad~\cite{huot2025agents}.
Complementarily, SCORE maintains long-term narrative coherence by tracking character and object states, constructing hierarchical episode summaries, and retrieving relevant prior episodes to detect and revise cross-episode inconsistencies~\cite{yi2026score}. 
More general figurative-language generation uses form-code variables to control transformations among literal, hyperbolic, idiomatic, sarcastic, metaphorical, and simile-based expressions~\cite{lai-nissim-2022-multi}. Across these tasks, generation is guided by an explicit account of the semantic relation or mechanism to be realized.

The second direction performs \textbf{affective and causal state modeling}, making users' emotions, underlying causes, beliefs, and support needs explicit before response generation. Sibyl infers causes, subsequent events, emotional states, and assistant intentions as future-aware commonsense~\cite{wang-etal-2025-sibyl}; ECTG models transitions between emotion--cause concepts across dialogue turns~\cite{icassp23emotioncausetransitiongraph}; and CAER introduces emotion--cause labels to improve causal grounding and response relevance~\cite{he2025ecc}. Psychology-inspired methods further organize support as a staged process. Chain-of-Empathy identifies emotions and their situational causes~\cite{lee2023chain}, Empathy-R1 reasons over context, causes, beliefs, communicative intents, and response strategies~\cite{yao2025empathy}, and HealMe decomposes cognitive reframing into distinguishing facts from feelings, considering alternative perspectives, and producing actionable support~\cite{xiao-etal-2024-healme}. Other systems represent the generator's communicative state directly: EmpHi learns latent empathetic intents such as acknowledging, consoling, questioning, and agreeing~\cite{chen-etal-2022-emphi}; ReflectDiffu maps emotion-contagion representations into actionable intents~\cite{yuan2025reflectdiffu}; and CAB jointly models cognition, affection, and behavior through commonsense paths, emotion variables, and dialogue acts~\cite{DASFAA23cab}. These approaches improve semantic grounding by linking a generated response to an explicit account of the user's situation and the response function it is intended to serve.

The third direction performs \textbf{interlocutor mental-state modeling}, representing the preferences, beliefs, intentions, attitudes, and anticipated reactions of conversational partners. This direction is particularly important for persuasion and negotiation, where an effective response depends on how the interlocutor is expected to interpret and react to it. PAN-DG conditions generation on argumentation profiles, preference profiles, buying styles, dialogue acts, and task knowledge, while related persona- and Theory-of-Mind-based systems incorporate personality and self--other awareness~\cite{priya-etal-2025-argue,zhao-etal-2023-dont,cai2024pecer}. PersuGPT generates candidate strategies and selects among them according to user intent and dialogue history~\cite{jin-etal-2024-persuading}. ToMAP predicts relevant counterclaims and estimates the opponent's agreement with each one, using the resulting counterclaim--attitude profile to guide persuasive responses~\cite{han2025tomap}. Studies on ChangeMyView similarly show that inferred intentions, emotions, and sentiments can support more targeted persuasive reasoning~\cite{amirizaniani2024llms}. Under information asymmetry, ToMMA distinguishes beliefs, desires, preventative behaviors, and generative behaviors before response construction~\cite{zhang2026persuasion}, whereas ToM-Agent tracks beliefs, desires, intentions, and confidence throughout the interaction~\cite{yang2025large}. These methods treat generation as adaptation to an inferred interlocutor state rather than response production from dialogue context alone.

Overall, mechanism-aligned semantic modeling introduces explicit representations of how high-level meanings are formed and communicated. Content-internal structures specify the semantic effect to be realized, affective and causal states explain why a response is appropriate, and interlocutor models determine how it should be adapted to a particular audience. Their main limitation is that errors in inferred mechanisms or mental states can propagate into generation, making the validity of intermediate representations as important as the quality of the final output.

\subsubsection{Synergistic Interactions among High-Level Semantics}
Beyond individual HLS categories, recent studies show that these abilities often interact with each other as semantic clues, control signals, or communicative mechanisms. 
Empathy and persuasion are the most explicit case: empathetic persuasive dialogue systems jointly optimize empathy and persuasiveness, while persuasive agents, negotiation agents, and conversational recommenders use emotional alignment, situational empathy, or empathetic expression to improve persuasive acceptance, decision making, and task success \cite{samad-etal-2022-empathetic, liu-etal-2025-dual}. % 10.1145/3706598.3713579  10.1145/3640457.3688133
Storytelling also functions as a mechanism for empathy and persuasion. 
Psychologically rich stories and narrative styles can evoke reader empathy, whereas narrative message formats change the effectiveness of persuasive communication \cite{harel-canada-etal-2024-measuring, shen-etal-2024-heart, Dai16122024}. 
Metaphor further supports storytelling and empathy by providing symbolic mappings that organize dream narration, scientific storytelling, and emotionally resonant empathetic responses \cite{10682437, lee-etal-2025-heart}. 
Humor interacts with persuasion and storytelling in a similar way: humorous explanation tones and stance-driven memes can influence decisions or persuasive message reception, while comic humor often depends on narrative sequencing, juxtaposition, or contradictory visual stories to construct punchlines \cite{10.1145/3706598.3713744, www24memecraft, ryan-etal-2025-humor, hu2024cracking}. 
Finally, multi-figurative generation suggests that metaphor and sarcasm can share non-literal properties and support cross-figurative transfer, indicating that different figurative HLS can be jointly modeled rather than treated as isolated labels \cite{lai-nissim-2022-multi}. 
Together, these works suggest that HLSI should not only recognize humor, empathy, metaphor, sarcasm, persuasion, and storytelling separately, but also model how one HLS mechanism reinforces or conditions another.

\subsubsection{Creative and Diverse Generation}

A central challenge in HLS generation is to increase creativity and diversity without sacrificing semantic fidelity. Creative outputs must avoid generic templates and dominant response patterns while remaining grounded in the intended semantic relation, communicative goal, or social context. Existing approaches introduce diversity at three main stages: inference-time exploration, representation and objective design, and data construction. They expand the candidate space through associative or multi-agent reasoning, represent multiple valid semantic realizations through latent variables and diversity-aware objectives, or broaden the underlying data distribution through retrieval and example selection.

One direction uses \textbf{associative and multi-agent exploration} to expand the conceptual and strategic search space before selecting a final output. Associative methods encourage models to move beyond literal or highly frequent continuations through weak associations, multi-hop reasoning, and divergent thought. CLoT frames creative humor as Leap-of-Thought rather than conventional Chain-of-Thought and uses weakly associated conditions with explorative self-refinement for multimodal Oogiri generation~\cite{a78fdd6694d54f5491f7393dfc7529e1,huang2025causality}. Its results suggest that weak associations produce more creative humorous responses than strongly associated conditions, whereas ordinary CoT provides limited benefits for LoT-style creativity. LoL similarly formulates humor as sparse multi-hop reasoning over knowledge graphs and employs instruction evolution, imaginative agents, and guided self-improvement to broaden the associative range~\cite{wang2025innovative}. GODBench's Ripple of Thought decomposes creative video commenting into ripple initiation, focalization, diffusion, interference, and final imprinting, explicitly encouraging divergent associations before selecting resonant comments~\cite{lei-etal-2025-godbench}. 
Multi-agent methods achieve a related goal by introducing diverse roles, perspectives, or interaction trajectories. For long-form narrative generation, StoryBox simulates character interactions within a dynamic sandbox, allowing events and plots to emerge organically before a Storyteller Agent organizes them into a coherent narrative~\cite{chen2026storybox}. 
Debate-to-write increases perspective diversity through persona-conditioned multi-agent debate~\cite{hu-etal-2025-debate}, while DEBATUNE generates and refines seed arguments through adversarial debate to produce diverse but stance-consistent statements~\cite{li-etal-2024-llms-speak}.
In persuasion, ToMAP predicts counterclaims over a cognitive graph to encourage new objections and preemptive arguments~\cite{han2025tomap}; DailyPersuasion expands seed concepts into long-tail topics, scenarios, and strategies~\cite{jin-etal-2024-persuading}; and multi-agent simulation frameworks monitor strategy repetition and reinforce speaker or cultural profiles to improve argument novelty~\cite{ma-etal-2025-communication,ma-etal-2025-enhancing}. Together, these methods increase diversity by making variation in associations, perspectives, and communicative strategies explicit during generation.

A second direction employs \textbf{diversity-aware representation and optimization} to prevent models from collapsing toward a small set of safe, frequent, or emotionally bland responses. Latent-variable methods represent multiple valid intents or affective states for the same context. EmpHi uses a CVAE with discrete empathetic intents to approximate the distribution of human response strategies~\cite{chen-etal-2022-emphi}, while CAB introduces separate latent variables for speaker and listener emotions to capture affective dependencies~\cite{DASFAA23cab}. ECC further combines a CVAE with an energy-based model and ODE sampling to compose emotion, cause, and commonsense factors in latent space~\cite{he2025ecc}. Other methods explicitly discourage generic outputs: SEEK and SERI apply frequency-aware cross-entropy, whereas NEC introduces negative examples from mismatched emotions, self-generated outputs, batch samples, and high-frequency statements~\cite{wang-etal-2022-empathetic,bi2023seri}. Similar objectives penalize repetition in polite persuasion~\cite{donate2savealife} or jointly optimize diversity, fluency, coherence, persona alignment, strategy consistency, and negotiation-phase agreement~\cite{priya-etal-2024-trip}. Preference-based optimization provides another mechanism for preserving unconventional but effective candidates. In cartoon captioning and storytelling image generation, DPO produces more diverse outputs than SFT, potentially because preference learning can exploit multiple acceptable human responses rather than imitate a single canonical output~\cite{zhang2024humor,song2025generating}. One proposed explanation is that maximum-likelihood SFT favors high-probability sequences and may therefore act as an ``entropy bottleneck,'' whereas reward-based optimization can reinforce less frequent reasoning paths when they achieve the intended semantic effect~\cite{zhang2026metaphorstar}. However, diversity-aware optimization must distinguish meaningful semantic variation from novelty, verbosity, or stylistic deviation alone.

A third direction introduces diversity through \textbf{data-level diversification}, broadening the semantic, lexical, and strategic distributions available during training or prompting. Synth-Empathy applies KCenterGreedy selection after quality filtering to construct a more diverse synthetic empathy dataset~\cite{liang2024synthempathy}. SYNTHESIZRR retrieves long-tail documents and grounds task inversion in external content, improving lexical and entity diversity while reducing repetition and popularity bias compared with few-shot generation~\cite{divekar-durrett-2024-synthesizrr}. Studies of synthetic sarcasm similarly find that grounding generation in naturally occurring examples improves faithfulness and topical diversity, whereas taxonomy-based generation may produce realistic-looking instances without reliably preserving the underlying semantic construct~\cite{veselovsky2023generating}. These findings show that output diversity depends not only on decoding and optimization, but also on the examples, retrieval sources, and sampling procedures from which models learn.

Overall, creative and diverse HLS generation requires coordinated intervention across inference, modeling, and data. Associative and multi-agent exploration expands the space of candidate ideas, diversity-aware representations and objectives preserve multiple meaningful realizations, and data-level diversification broadens the distribution from which generation begins. The central challenge is to ensure that increased diversity reflects meaningful semantic and strategic variation rather than randomness or loss of communicative intent.

\subsubsection{Multimodal Semantic Integration and Generation}

HLS generation often requires models to combine textual content with visual, acoustic, temporal, and paralinguistic cues, since humor, metaphor, sarcasm, empathy, and persuasion may emerge from interactions across modalities. Existing methods address this challenge through three complementary strategies: converting heterogeneous inputs into explicit semantic representations, grounding and selectively integrating fine-grained multimodal clues, and using multimodal signals to control the semantic and expressive realization of outputs.

The first strategy adopts \textbf{representation-first multimodal generation}, converting heterogeneous sensory inputs into textual or symbolic evidence before producing the final output. Rather than requiring a generator to interpret raw modalities directly, these methods extract captions, OCR text, speaker mappings, character identities, object descriptions, emotions, or communicative intentions that can be inspected and transformed by language models. Sticker-based dialogue summarization, for example, uses GPT-4V-generated descriptions of sticker emotions and intentions to support textual summarization, while manga-to-prose generation decomposes visual narratives into panel structure, OCR text, speaker associations, image captions, and character grounding before producing prose~\cite{10.1145/3664647.3680978,sachdeva2025panels}. GODBench similarly represents short videos through content descriptions, cultural context, emotional resonance, and divergent associations before generating creative comments~\cite{lei-etal-2025-godbench}. These approaches make multimodal evidence more accessible and interpretable, although textual abstraction may discard fine-grained visual, temporal, or acoustic information.

The second strategy performs \textbf{fine-grained cross-modal grounding and selective fusion}. Its central challenge is to align generated content with relevant visual regions, objects, sub-images, acoustic patterns, or modality-specific units while suppressing incomplete or misleading signals. In multimodal humor, HUMORCHAIN, VCDHG, and IRCoT decompose visual context, identify incongruity, or apply theory-guided reasoning before generating image-conditioned humorous content~\cite{zhang2025humorchain,11209232,tanaka-etal-2024-content}. XMeCap further grounds meme captions through sub-image detection and token--region alignment based on cross-modal attention and similarity objectives~\cite{mm24xmecap}. For visual metaphor generation, LLM--diffusion pipelines first elaborate implicit metaphorical meanings and visual objects through language before image synthesis, while GOME modifies diffusion cross-attention to associate metaphorical attributes with the correct objects~\cite{chakrabarty-etal-2023-spy,zhang-etal-2024-gome}. In empathy-oriented generation, ERGM combines visual and acoustic features with cross-modal attention and caption guidance, whereas MERIA separates shared emotional information, modality-private information, and identity features before applying uncertainty-gated fusion for Chain-of-Empathy reasoning~\cite{11129652,10.1145/3746027.3762031}. Collectively, these methods show that multimodal integration requires both precise alignment and adaptive selection rather than uniform fusion of all available signals.

The third strategy \textbf{uses multimodal signals to control semantic and expressive realization}. These methods infer high-level states such as emotion, personality, empathy, social beliefs, sarcasm, or speaking style and use them as control variables for generating text, speech, facial motion, or other responses. E3RG decomposes multimodal empathetic response generation into empathy understanding, memory retrieval, and response generation, using predicted emotions to guide both speech and talking-head outputs~\cite{lin2025e3rg}. MultiMind integrates text, facial expressions, and vocal tone into belief modeling for strategic communication, while personality-aware generation combines visual--textual encoding with MBTI-based personality control~\cite{zhang2025multimind,wu-etal-2025-traits}. Speech-oriented systems further demonstrate that HLS must be realized acoustically as well as lexically. Sarcastic speech synthesis employs sarcasm representations or retrieved prosodic exemplars~\cite{li25b_ssw,li2025making}, while OSUM-EChat reasons over speaker characteristics, emotions, and sound events~\cite{geng2025osum}. Emotion Omni, Chain-Talker, GPT-Talker, and ParalinGPT incorporate emotional speech features, acoustic history, style tokens, speech embeddings, or sentiment labels to produce more expressive dialogue~\cite{wang2025empathyomnienablingempathetic,hu-etal-2025-chain,liu2024generative,lin2024paralinguistics}. The effectiveness of such control depends on whether the inferred states are accurate and whether they are consistently realized across output modalities.

Overall, multimodal HLS generation is not simply multimodal perception followed by text generation. It involves abstracting heterogeneous evidence, grounding semantic content in fine-grained multimodal clues, selectively resolving conflicting signals, and realizing the intended high-level meaning through appropriate verbal and nonverbal forms.

\subsubsection{Strategy-Guided Controllable Generation}

Controllable HLS generation exposes explicit rhetorical, pragmatic, or communicative strategies, allowing models to specify not only what to generate but also how the intended semantic effect should be realized. Existing methods mainly provide control through global strategy selection or fine-grained realization.

One direction uses \textbf{global strategy-level control}. In humor generation, VCDHG annotates rhetorical devices and trains MLLMs through strategy identification and rewriting~\cite{11209232}. HUMORCHAIN derives generation strategies from humor and psychological theories, while TalkFunny uses chain-of-humor annotations, humor mind maps, and rewriting to control humorous content and affective style~\cite{zhang2025humorchain,Chen_Yuan_Liu_Liu_Guan_Guo_Peng_Liu_Li_Xiao_2024}. Persuasion systems similarly select context-appropriate strategies through special tokens~\cite{song-wang-2024-like}, adapt politeness according to user resistance~\cite{donate2savealife}, or choose among credibility-aware strategies such as logical appeal, emotional appeal, evidence, and social proof~\cite{qin-etal-2024-beyond}. Making strategies explicit improves interpretability and alignment with task-specific communicative goals.

A second direction provides \textbf{fine-grained control over strategy realization}. For empathetic generation, EmpHi controls responses through latent intents and intent keywords, CAB uses dialogue acts, and APTNESS predicts support strategies before generation~\cite{chen-etal-2022-emphi,DASFAA23cab,10.1145/3627673.3679687}. DIFFUSEMP further uses communication mechanisms, intents, semantic frames, and control-range masking to determine which tokens each control signal influences~\cite{bi-etal-2023-diffusemp}. In speech generation, sarcasm embeddings, retrieved prosodic exemplars, and intonation tokens control acoustic realization~\cite{li25b_ssw,li2025making,aaaiProsodyFM}. Multimodal systems also condition meme generation on social causes, stances, persuasion techniques, image structures, or emotion categories~\cite{www24memecraft,mm24xmecap}. These methods extend controllability from global strategy choice to token-level, prosodic, and visual realization.

Overall, explicit controls improve predictability and task fit, but overly rigid strategies may reduce creativity and naturalness~\cite{zhang2025humorchain,a78fdd6694d54f5491f7393dfc7529e1}. Effective control should therefore preserve flexibility in how the intended semantic function is expressed.

\subsubsection{External Knowledge Integration}

HLS generation often requires factual, cultural, psychological, linguistic, or context-specific knowledge that is not fully contained in the immediate input. Existing methods integrate such knowledge in three main ways: retrieving external evidence or persistent task-specific memory at inference time, organizing generation through structured knowledge scaffolds, or internalizing external knowledge through training.

One direction uses \textbf{retrieval-augmented grounding} to condition generation on external exemplars, knowledge bases, domain resources, or persistent contextual memory rather than relying solely on parametric memory.
PERGM retrieves exemplary empathetic responses based on persona--context similarity and converts them into generation prompts~\cite{huang2024generating}, while coTherapist retrieves evidence-based passages from psychotherapy manuals and clinical guidelines~\cite{www26coTherapist}. KEMI combines COMET-expanded commonsense queries with subgraph retrieval over a mental-health knowledge graph, using retrieved cases to improve support-strategy prediction and response specificity~\cite{deng-etal-2023-knowledge}. Retrieval also supports other HLS tasks: 
For long-form narrative generation, SCORE treats prior episodes and hierarchical summaries as retrievable story memory, combining lexical and semantic retrieval to recover relevant events and maintain cross-episode consistency~\cite{yi2026score}. 
AMBIPUN uses sense definitions, reverse dictionaries, and external contexts to construct ambiguous pun settings~\cite{mittal-etal-2022-ambipun}; PC-CRS grounds persuasive explanations in item information and critiques unsupported claims against source facts~\cite{qin-etal-2024-beyond}; and sarcasm-aware speech synthesis retrieves sarcastic utterances as prosodic references~\cite{li25b_ssw}. Retrieval improves grounding and knowledge updateability, although its effectiveness depends on evidence relevance and reliability.

A second direction uses \textbf{structured knowledge as generation scaffolds}, converting external commonsense, linguistic, causal, or theoretical knowledge into intermediate structures that guide surface realization. CASE constructs cognition and emotion graphs from COMET and ConceptNet and aligns them to support empathetic response generation~\cite{zhou-etal-2023-case}. ECTG instead derives an Emotion Cause Transition Graph from dialogue data to model how causal concepts move from context to response~\cite{icassp23emotioncausetransitiongraph}. In pun generation, WordNet definitions, word-sense disambiguation, non-pun corpora, and humor principles provide explicit constraints on ambiguity, distinctiveness, and surprise~\cite{tian-etal-2022-unified}. GOME structures visual metaphor generation through rhetorical knowledge about tenor, vehicle, and grounding~\cite{zhang-etal-2024-gome}, while persuasion research uses a taxonomy of 40 techniques derived from social-science literature to guide strategic paraphrasing~\cite{zeng-etal-2024-johnny}. Such scaffolds improve interpretability and control, but remain limited by the coverage and correctness of the underlying resources.

A third direction performs \textbf{parametric knowledge internalization}, incorporating external domain or theory-based knowledge into model parameters during training. SYNTHEMPATHY uses psychotherapy-informed prompts to construct explanation--response pairs for fine-tuning empathetic response models~\cite{chen2025synthempathy}. Psyche-R1 integrates psychological expertise from textbooks, question banks, support platforms, and external datasets through synthetic instruction data, SFT, and GRPO~\cite{dai-etal-2026-psyche}. In persuasion, DailyPersuasion derives scenarios and strategies from sociological and psychological principles, which PersuGPT internalizes through strategy reasoning and preference optimization~\cite{jin-etal-2024-persuading}. Parametric internalization enables efficient generation without live retrieval, but makes knowledge provenance less transparent and subsequent updates more difficult.

Overall, retrieval provides dynamic access to external evidence, structured scaffolds organize knowledge into interpretable generation mechanisms, and training-time fusion internalizes it into model parameters. These approaches are complementary, but all depend on the quality, relevance, and cultural validity of the external knowledge being introduced.

\subsubsection{Personalized and User-Adaptive Generation}

HLS effects such as empathy, humor, persuasion, and stance alignment are often perceived differently across users. Models therefore incorporate personas, preferences, emotional states, dialogue histories, social roles, or interaction contexts rather than generating from task inputs alone. Existing approaches mainly differ in whether they condition on relatively stable user profiles or dynamically update user states during interaction.

One direction uses \textbf{profile-conditioned personalization}, adapting generation to relatively stable user traits, histories, or preferences. In empathetic response generation, PERGM retrieves persona--context-matched exemplars and combines them with emotion and dialogue-act representations, while PECER and personality-aware multimodal models infer personality traits from dialogue histories or multimodal cues and use them as generation controls~\cite{huang2024generating,cai2024pecer,wu-etal-2025-traits}. Personalized headline generation similarly extracts entity and syntactic preferences from a user's historical headlines and incorporates them through pointer mechanisms or prefix tuning~\cite{getthepunchline}. In persuasive dialogue, PersuaSim-Reinforced uses speaker profiles and native-language prompting to improve informativeness, argument novelty, stance stability, and strategy diversity across cultural settings~\cite{ma-etal-2025-enhancing}. These methods reduce generic or majority-aligned responses by conditioning generation on persistent individual or group characteristics.

A second direction performs \textbf{dynamic user-state adaptation}, estimating evolving emotions, intentions, resistance, constraints, or mental states and adjusting generation accordingly. In emotional support, self-evolution frameworks infer user profiles, situations, and emotional states to construct preference pairs for iterative optimization, while KardiaBench and Kardia-R1 ground multi-turn support in real user profiles and optimize persona consistency, emotional reasoning, and empathy~\cite{ye-etal-2025-generic,10.1145/3774904.3793022}. In persuasion and negotiation, PersuGPT estimates the downstream effects of candidate responses through a simulated user model~\cite{jin-etal-2024-persuading}; politeness-aware dialogue and ToMAP adapt strategies according to resistance, politeness intensity, and attitudes toward counterclaims~\cite{donate2savealife,han2025tomap}; and tourism negotiation systems model traveler personas, preferences, intents, buying styles, and negotiation phases~\cite{priya-etal-2024-trip,priya-etal-2025-argue}. MADeN similarly adjusts debt-collection strategies according to debtor types, while ToM-Agent tracks beliefs, desires, and intentions to support socially adaptive responses~\cite{wang2025debt,yang2025large}. Unlike static personalization, these methods treat the user model as an evolving state that must be revised throughout the interaction.

Overall, profile-conditioned methods capture persistent user differences, whereas dynamic adaptation responds to changing states and interaction trajectories. However, stronger personalization may undermine factuality, safety, and fairness when user profiles are inaccurate or problematic preferences are amplified. User-adaptive generation should therefore combine personalization with explicit safeguards and evaluation criteria to ensure that adaptation remains reliable and socially appropriate.

\subsubsection{Data-Efficient Learning}

HLS generation often suffers from limited supervision because high-quality examples are subjective, culturally situated, and expensive to annotate. Existing methods address this problem by introducing structural priors, exploiting indirect supervision, and adapting pretrained models with data-efficient training procedures. These approaches seek to make each example more informative and reduce dependence on large task-specific datasets.

One direction \textbf{introduces task structure and prior knowledge as inductive biases}. In humor generation, AMBIPUN avoids direct training on large pun corpora by retrieving sense-specific context words and using them to guide a T5 model trained on general non-humorous text~\cite{mittal-etal-2022-ambipun}. Metaphor generation similarly decomposes figurative meaning into explicit semantic components or explanation-based formulations, making implicit relations easier to learn from limited data~\cite{feng-ma-2022-better,chakrabarty-etal-2022-flute}. For persuasion, ToMMA incorporates causal Theory-of-Mind constraints so that generation is guided by mental-state reasoning rather than surface dialogue imitation alone~\cite{zhang2026persuasion}. From a data-efficiency perspective, these methods replace part of the missing supervision with linguistic, pragmatic, or reasoning structure.

A second direction \textbf{learns from indirect supervision}, including pseudo-labels, automatically derived annotations, teacher feedback, preference comparisons, and retrieved exemplars. Nominal metaphor generation combines limited labeled data with unlabeled corpora through self-training and weights pseudo-labeled examples according to their estimated metaphor probability~\cite{li-etal-2022-nominal}. In humor generation, a unified pun framework combines automatic labels with a small set of human-annotated disagreements, while humor distillation transfers paraphrasing ability from a large teacher to a smaller model through generated examples and critique feedback~\cite{tian-etal-2022-unified,ravi-etal-2024-small}. Politeness-aware persuasion derives training signals from automatic politeness annotation and paraphrase-based transfer pairs~\cite{donate2savealife}, whereas self-evolving empathy systems infer user profiles, situations, and emotions to construct preference pairs for iterative DPO~\cite{ye-etal-2025-generic}. These approaches transform weak or implicit signals into usable supervision, although errors in automatic labels or teacher feedback may also be propagated during training.

A third direction uses \textbf{staged and parameter-efficient adaptation} to transfer pretrained capabilities to small HLS datasets. Context-situated pun generation combines broad pretraining, intermediate fine-tuning, and limited task-specific supervision to improve pun incorporation and generation success~\cite{sun2022context}. Sarcasm-aware speech generation similarly progresses from general speech pretraining to conversational speech and then to small sarcasm-labeled subsets~\cite{li25b_ssw}. Related work combines LoRA adaptation with retrieved prosodic exemplars, reducing reliance on directly annotated sarcastic speech~\cite{li2025making}. PGCL further applies LoRA-based preference learning and staged curriculum alignment to small-scale pun datasets~\cite{chen-etal-2024-u}. These methods reduce training cost and overfitting, but their effectiveness depends on whether the pretrained model already contains transferable semantic and pragmatic knowledge.

Overall, data-efficient HLS generation relies not only on obtaining more examples, but also on extracting more supervision from available data, introducing appropriate inductive biases, and reusing pretrained capabilities through efficient adaptation.

\subsection{Evaluation}
\label{sec:hlsg_eval}

HLS generation is inherently open-ended, and a single input may admit multiple valid outputs with different semantic realizations. Evaluation must therefore consider not only reference similarity, but also subjective quality, communicative effects, control adherence, multimodal consistency, creativity, and downstream utility. Existing methods can be organized into six complementary categories.

\subsubsection{Reference-Based and Semantic-Similarity Evaluation}

Reference-based evaluation compares generated outputs with human-written references at the lexical or semantic level. Lexical metrics such as BLEU, ROUGE, METEOR, and CIDEr measure n-gram or sequence overlap, whereas BERTScore, BLEURT, BARTScore, COMET, and embedding similarity capture broader semantic correspondence. These metrics are efficient and reproducible and are widely used in empathetic response generation, figurative-language explanation, multimodal summarization, and metaphor captioning. CASE and Sibyl, for example, combine perplexity and reference-based metrics with Distinct scores to evaluate response quality and diversity~\cite{zhou-etal-2023-case,wang-etal-2025-sibyl}. However, reference similarity may correlate weakly with human judgments when outputs are creative, figurative, culturally situated, or pragmatically effective in multiple valid ways~\cite{wang-etal-2025-proverbs}. Reference-based metrics are therefore most informative when used alongside effect- or judgment-based evaluation.

\subsubsection{Judgment-Based Evaluation}

Many important qualities of HLS outputs, such as empathy, humor, creativity, persuasiveness, and naturalness, cannot be reliably captured by reference overlap or a single task-specific metric. 
Judgment-based quality evaluation therefore relies on human or model evaluators to assess outputs holistically through multidimensional ratings, comparative preferences, or predefined rubrics. 

\textbf{Human-centered evaluation} directly asks annotators, experts, or users to assess dimensions such as meaningfulness, empathy, humor, creativity, persuasiveness, naturalness, and emotional appropriateness. Common protocols include Likert ratings, pairwise preferences, win--tie--lose comparisons, expert rankings, blind review, and user studies. Empathetic responses are commonly compared in terms of empathy, relevance, and fluency~\cite{huang2024generating}, while humor generation relies on human or expert judgments of funniness, creativity, and contextual appropriateness~\cite{zhang2025humorchain,mittal-etal-2022-ambipun,zhang2024humor}. Visual-metaphor studies ask artists or participants to evaluate meaning preservation and emotional expressiveness~\cite{chakrabarty-etal-2023-spy}, whereas persuasion systems assess perceived persuasiveness, politeness, coherence, and preference~\cite{breum2024persuasive,donate2savealife}. Human evaluation captures subjective semantic effects directly, but is costly and sensitive to annotator background and protocol design.

\textbf{Model-based evaluation} uses LLMs, reward models, critics, or trained discriminators as scalable judges. These evaluators may score outputs with predefined rubrics, compare candidates, detect whether a target semantic function is present, or provide rewards during optimization. GPT-style judges and psychology-grounded reward models have been used to assess empathy, supportiveness, persona consistency, and safety~\cite{10.1145/3627673.3679687,wang-etal-2025-sibyl,chen-etal-2024-emotionqueen,wang2026permpsychologygroundedempatheticreward}. Humor systems use LLM critics or discriminators to rank and filter generated jokes or captions~\cite{zhang2025humorchain,hwang-etal-2025-bottlehumor,ravi-etal-2024-small}, while persuasion and social-media generation employ LLM judges for response quality and pairwise win rates~\cite{jin-etal-2024-persuading,zhang-etal-2024-somelvlm}. 
For narrative generation, STORYRMB benchmarks whether reward models can recover human story preferences across coherence, creativity, characterization, fluency, and relevance, while STORYREWARD provides a story-specific evaluator trained on large-scale preference data~\cite{xia2026storyalign}.
Because model judges may exhibit position, length, cultural, or model-specific biases and may fail to capture domain-specific preferences, their scores should be validated against human judgments rather than treated as definitive.

\subsubsection{Semantic Attribute and Control Evaluation}

This evaluation examines whether generated outputs exhibit specified semantic attributes and satisfy explicit control conditions. It uses task-specific classifiers, structured checks, and attribute-level metrics to diagnose which properties are successfully realized.

\textbf{Semantic-attribute evaluation} measures properties such as empathy level, communicative intent, pun structure, sarcasm recognizability, persuasion strategy, and narrative coherence. Emp-F1 evaluates agreement with target empathy levels, while EmpHi and ReflectDiffu assess emotion and intent distributions~\cite{ma2025emprl,chen-etal-2022-emphi,yuan2025reflectdiffu}. Humor and pun systems verify structural validity, pun-word incorporation, safety, and other phenomenon-specific attributes~\cite{chen-etal-2024-u,sun2022context,aaai26humorreject}. Sarcastic speech is evaluated with detector-based recognition scores~\cite{li2025making}, while persuasive generation considers strategy accuracy, politeness, credibility, and rhetorical mode~\cite{donate2savealife,breum2024persuasive}. 
For long-form narratives, SCORE evaluates cross-episode coherence on the NCI-2.0 benchmark, emotional consistency with the EASM metric, and the occurrence of hallucinated story elements~\cite{yi2026score}. 

\textbf{Control-condition evaluation} determines whether outputs follow prescribed conditions such as persona, emotion, stance, culture, user preference, dialogue act, or figurative style. Empathetic systems evaluate persona consistency and preference alignment~\cite{huang2024generating,ye-etal-2025-generic}. Persuasion and negotiation systems assess adherence to personas, cultural profiles, negotiation phases, and viewer characteristics~\cite{priya-etal-2024-trip,ma-etal-2025-enhancing,kim-etal-2025-pvp}. Figurative generation jointly measures control strength, semantic preservation, and fluency~\cite{lai-nissim-2022-multi}.

\subsubsection{Multimodal-Consistency Evaluation}

Multimodal HLS generation must be evaluated both as an output and as an interpretation of visual, acoustic, video, or other non-textual evidence. Metrics such as CLIPScore, RefCLIPScore, $\Delta$-CLIPScore, VQA-based scores, emotion alignment, speaker similarity, UTMOS, NISQA, and ASR-WER are commonly combined with human judgments of relevance, faithfulness, naturalness, and cross-modal coherence. Humor and meme-generation studies assess whether captions or explanations reflect visual incongruity, stance, meme semantics, and relevant sub-images~\cite{11209232,www24memecraft,hwang-shwartz-2023-memecap}. Spoken-empathy and expressive-dialogue systems evaluate whether speech, prosody, emotion, and paralinguistic cues consistently realize the intended meaning~\cite{geng2025osum,wang2025empathyomnienablingempathetic,hu-etal-2025-chain}. This evaluation helps distinguish failures of multimodal grounding from failures in surface realization.

\subsubsection{Diversity and Creativity Evaluation}

Because HLS generation is open-ended, evaluation should also measure whether outputs are diverse, original, and distributionally faithful. Distinct-$n$ and Self-BLEU capture lexical repetition, while semantic diversity, entity entropy, MAUVE, originality overlap, perspective diversity, and creativity scores assess broader variation in meaning or viewpoint. In empathetic dialogue, diversity metrics identify generic and high-frequency responses~\cite{he2025ecc}. Humor and pun studies additionally evaluate originality, ambiguity, surprise, and creative deviation~\cite{xu-etal-2024-good,tian-etal-2022-unified,alnajjar-etal-2022-laugh}. Synthetic-data and multi-agent generation studies examine whether outputs avoid repetition, popularity bias, and narrow perspectives while preserving distributional realism~\cite{divekar-durrett-2024-synthesizrr,hu-etal-2025-debate}. Since lexical variation does not necessarily imply semantic creativity, automatic diversity scores should be complemented with human or model-based judgments.

\subsubsection{Extrinsic and Downstream Evaluation}

Extrinsic evaluation measures whether generated outputs achieve practical goals in downstream tasks or real interactions. Persuasion studies use donation rates, attitude shifts, agreement changes, negotiation success, and opinion-change probabilities~\cite{donate2savealife,han2025tomap,breum2024persuasive,chi24audiencedepolarisation}. Visual-metaphor generation can be evaluated through improvements in visual entailment or retrieval~\cite{chakrabarty-etal-2023-spy,zhang-etal-2024-gome}. Mental-health and empathetic-dialogue systems assess perceived support, safety, clinical usefulness, and expert-rated quality~\cite{bn2025pursuit,www26coTherapist}. Although extrinsic evaluation provides the strongest evidence of real-world utility, observed outcomes may also depend on users, contexts, and downstream systems beyond the generated content itself.

%% file: 05_applications.tex
\section{Applications}
\label{sec:applications}

\subsection{High-Level Semantics and AI Safety}

Safety remains an indispensable concern in AI development, and HLS is deeply connected to AI safety, with broad potential for safety-critical applications. 

Safety-oriented applications highlight the necessity of HLS understanding beyond the literal meaning captured by BLS. 
Many harmful expressions are not lexically explicit, but are encoded through implication, irony, metaphor, argot, presupposition, or conversational context. 
For implicit hate speech detection, CoSyn incorporates user history and conversational structure to infer harmful intent even when overt toxic words are absent, while Cracking the Code treats implicit abuse as a problem of decoding rhetorical devices such as irony, metaphor, and coded language \cite{ghosh-etal-2023-cosyn, wei-etal-2025-cracking}. 
Causality-guided and pragmatic-reasoning methods further show that safety models need to separate genuine harmful intent from superficial stylistic cues, improving robustness to cross-style or inference-intensive toxic language \cite{10.1145/3774904.3792669, chen-wang-2025-pragmatic}. 

Metaphor is especially important in safety because it can function both as a defensive signal and as an adversarial attack surface. 
On the defensive side, metaphor-aware models are used to identify misogyny, implicit hate, or offensive meanings in memes and figurative expressions where the target is not literally named \cite{mia-etal-2025-banmime, zeng-etal-2025-sheeps}. 
On the adversarial side, jailbreak studies show that attackers can exploit metaphorical mappings to transform benign-looking prompts into harmful instructions. 
For example, AVATAR uses adversarial metaphors to map harmless entities onto toxic ones, while metaphor-based jailbreaks against text-to-image models show that figurative prompts can bypass filters by hiding sensitive semantics behind symbolic language \cite{yan-etal-2025-benign, zhang2025metaphor}. 
These works indicate that robust safety requires reasoning about latent source-target mappings and communicative intent rather than only detecting explicit unsafe tokens. 

Humor and sarcasm create another major safety challenge because harmful meanings can be disguised as jokes, satire, dark humor, or playful exaggeration. 
Engagement Undermines Safety shows that optimizing LLMs for funniness can amplify stereotypes and toxicity, suggesting that humor generation may learn harmful cues as shortcuts for engagement \cite{dogra-etal-2026-engagement}. 
Dark-humor and meme-based safety benchmarks further require models to identify hidden targets, intensity, and social abuse beneath humorous framing, rather than simply deciding whether content appears funny or offensive on the surface \cite{kasu2025d, lin2026goat}. 
At the same time, humor can also be used constructively: HumorReject uses humorous indirect refusals to defuse harmful requests and reduce the brittleness of explicit refusal prefixes \cite{aaai26humorreject}. 
Thus, humor and sarcasm are not peripheral creative skills in safety; they directly affect both the generation of harmful content and the design of safer refusal strategies.

Safety work also increasingly studies multimodal and culturally situated harmful content, where social meaning emerges from the interaction between text, image, target identity, and background knowledge.
Datasets such as MIMOSA, BanMiMe, and BHM focus on low-resource or culturally specific meme moderation, requiring models to understand aggression, misogyny, hate targets, metaphor, humor, and code-mixed language across modalities~\cite{ahsan-etal-2024-multimodal, mia-etal-2025-banmime, hossain-etal-2024-deciphering}.
Other work extends this direction to propaganda, hateful videos, multimodal social abuse, and text-to-image jailbreaks, where harmful intent may depend on visual symbols, cultural references, cross-modal contradictions, or the composition of individually safe modalities~\cite{10.1007/978-981-96-0576-7_28, jing-etal-2025-hvguard, lin2026goat, liu-etal-2025-multimodal-pragmatic}.
Text-to-image jailbreak work further shows a form of semantic upgrading, where unsafe meaning emerges compositionally from visual text and image content that appear safe in isolation, exposing the limits of single-modality keyword, prompt, and NSFW filters for diffusion-model safety~\cite{liu-etal-2025-multimodal-pragmatic}. 
Multi-agent debate frameworks such as MV-Debate further use multiple reasoning perspectives to analyze surface clues, implicit meanings, modality conflicts, and social context, making safety judgments more interpretable and robust~\cite{lu2025mv}.

Finally, social and affective capabilities can serve both as safety mechanisms and as sources of new risks. 
Lahnala et al.~\cite{lahnala-etal-2022-mitigating} show that empathy-scored data can mitigate toxic degeneration with substantially fewer fine-tuning examples, with cognitive components such as interpretation and exploration proving more effective than emotional empathy. 
However, highly empathetic language may also produce affective hallucination, causing users to perceive illusory emotional connections or become overdependent on the system~\cite{kim-etal-2026-kind}. 
Persuasion-based jailbreak studies similarly show that attackers can exploit social pressure, emotional framing, role-play, or human-like persuasion to induce models to violate safety constraints~\cite{zeng-etal-2024-johnny}. 
Together, these studies suggest that HLS is double-edged in AI safety: the same capabilities that can support safer generation may also create new vulnerabilities or be exploited by attackers.

\subsection{Broader Application Domains}

\textbf{Mental health and affective support} use HLS to move from surface-level emotional mirroring to deeper interpretation of users' psychological states. Empathy is used to infer users' emotions, causes, needs, and appropriate support strategies. Chain of Empathy guides LLMs through psychotherapy-inspired reasoning before generating responses~\cite{lee2023chain}. 
HealMe applies cognitive reframing to help users reinterpret negative thoughts, showing how semantic understanding of emotion, belief, and cause can support psychotherapy-oriented dialogue \cite{xiao-etal-2024-healme}. 
Metaphor and storytelling also appear in this setting because users often express distress through personal narratives, symbolic images, or memes; figurative mental-health meme classification uses metaphor and commonsense knowledge to infer anxiety- or depression-related meanings that are not literally stated \cite{mazhar2025figurative}. 

\textbf{Social agents and human-like interaction} apply empathy, humor, persona, and persuasion to make AI systems behave as socially situated interlocutors rather than generic assistants. 
EmpathyAgent evaluates whether embodied agents can understand emotional needs and perform appropriate supportive actions, extending empathy from language to situated behavior \cite{chen2025empathyagent}. Humor-aware social robots use context, user profiles, and emotional states to generate more engaging interactions \cite{11127367}. 
Persona-related work such as Open Character Training controls stable traits like warmth, humor, and care through Constitutional AI and synthetic introspective data, while PrefPalette models how different communities value attributes such as empathy, humor, and directness \cite{maiya2025open, li2025prefpalette}. 

\textbf{Negotiation and decision support} use HLS to model preferences, beliefs, intentions, trade-offs, and strategic communication. NegotiationToM evaluates whether models can track beliefs, desires, and intentions in negotiation, showing that persuasive and negotiative dialogue requires mental-state reasoning rather than only fluent argument generation \cite{chan2024negotiationtom}. ToMAP trains opponent-aware persuaders that adjust their strategies according to interlocutors' attitudes and counterclaims \cite{han2025tomap}. In applied decision settings, travel negotiation, privacy-setting negotiation, task allocation, and conversational recommendation systems use persuasion and preference modeling to help users reach agreements or make choices under competing constraints \cite{priya-etal-2024-trip, 10.1145/3613904.3642897, Zahedi_Sengupta_Kambhampati_2024, qin-etal-2024-beyond}.

\textbf{Creative content and entertainment} use storytelling, humor, sarcasm, and metaphor to generate or interpret jokes, comics, memes, satire, and narratives. From Panels to Prose converts manga panels into literary narratives, applying storytelling semantics to track panels, characters, dialogue, and narrative flow \cite{sachdeva2025panels}. STRIPCIPHER evaluates whether multimodal models can recover implicit narrative order in silent comic strips, while FlawedFictions tests whether models can detect plot holes and maintain long-range story consistency \cite{wang2025beyond, ahuja2025finding}. Humor-focused work such as YESBUT and HUMORCHAIN tests whether models can understand contradictory comic humor or generate interpretable multimodal humor through incongruity reasoning \cite{liang2025yes, zhang2025humorchain}. 

\textbf{Cultural and artistic understanding} uses metaphor, humor, storytelling, and implicit social reasoning to interpret symbolic expression and culturally grounded meaning. The Chinese Pun Rebus Art dataset evaluates whether models can understand symbolic associations grounded in Chinese cultural knowledge beyond literal image recognition \cite{zhang2025creating}. The Language of Infographics studies conceptual metaphor in scientific storytelling, showing how visual metaphors help communicate abstract scientific ideas through more accessible narrative forms \cite{10682437}. Art-context evaluation connects Theory of Mind and critique generation, requiring models to infer artistic intention, audience interpretation, and culturally situated meaning \cite{arita2025assessing}. These works show that HLS is important not only for entertainment but also for cultural interpretation and knowledge communication.

%% file: 06_future.tex
\section{Challenges and Future Directions}
\label{sec:future_directions}

Although HLS has been extensively studied and substantial progress has been made, current models remain far from achieving human-expert-level HLSI. Accordingly, this section discusses the limitations of existing HLS research and outlines several promising directions for future exploration.

\subsection{Scaling Up and Utilizing High-Level Semantic Data}

HLS data are characterized by complex reasoning processes and stringent quality requirements. 
Their construction often requires substantial domain knowledge, making them difficult to generate at scale and consequently scarce. 
More importantly, existing data often provide only outcome-level annotations, while fine-grained supervision signals for intermediate reasoning processes and the mechanisms underlying HLS remain particularly limited. 
An important research direction is therefore to develop effective methods for mining or synthesizing both high-quality HLS data and reasoning-level supervision signals.

HLS data have broad application potential. 
Beyond improving task-specific performance, they can enhance models’ general capabilities by exposing them to abstract reasoning, implicit meaning construction, and complex contextual interpretation. 
For instance, COIG-CQIA~\cite{bai2025coig} shows that humorous and absurd content from RuoZhiBa incorporates logical structures, cognitive and linguistic traps, jokes, riddles, and artistic or abstract rhetoric. Fine-tuning on such content challenges models’ multi-hop reasoning, enhances their understanding of Chinese, and improves complex logical reasoning. 
Similarly, MetaphorStar~\cite{zhang2026metaphorstar} demonstrates that training on metaphorical images requiring multi-hop reasoning, cultural context, and ToM capabilities improves general comprehension, particularly complex visual reasoning.
Thus, jointly exploring the scalable construction and effective utilization of HLS data is a promising direction for advancing HLSI capabilities.

\subsection{Assessment of Abstract High-Level Semantics}

Due to the abstract and subjective nature, HLS assessment remains a major challenge. 
Nevertheless, reliable semantic assessment is essential for data filtering, task evaluation, and controllable generation. 
As mentioned earlier, HLS can be assessed from multiple complementary perspectives, including intrinsic semantic assessment and receiver reaction modeling. 

From the perspective of intrinsic semantic assessment, HLS can be measured in terms of the size or complexity of its semantic content~\cite{scott2019glasgow, Song_Pang_Tang_Wu_Zhu_2025}, but its inherent abstractness makes such measurement particularly challenging, and systematic research remains limited. 
Alternatively, HLS can be assessed through its semantic effects, as reflected in receivers’ emotional, cognitive, or behavioral responses~\cite{buechel2018modeling,ma2025emprl,alnajjar-etal-2022-laugh,bai2021m2p2}. 
However, accurately modeling these reactions remains challenging. 
Beyond task evaluation, modeling semantic receivers’ reactions is important for applications such as role-playing and user simulation, where systems must predict not only what an expression means but also how different receivers perceive, interpret, and respond to it.

\subsection{Perception of Semantic Clues}

Understanding and generating HLS, particularly in multimodal settings, require models to first identify and interpret relevant semantic clues before performing higher-level reasoning.
Existing benchmarks accordingly distinguish perceptual capabilities from cognitive reasoning~\cite{fu2023mme,song-etal-2025-cognitive}, while evaluations of integrated multimodal capabilities show that complex task performance depends on the composition of more basic skills such as recognition, spatial awareness, OCR, and knowledge grounding~\cite{yu2024mmvet}.
Moreover, deficiencies in basic visual representations can propagate into incorrect reasoning and hallucinated explanations, even when models possess strong language-based reasoning capabilities~\cite{Tong_2024_CVPR,kanade-ganu-2026-see}.
Similar dependencies arise in specific HLS tasks: for example, sarcasm and social-intelligence understanding benefit from accurately capturing and integrating linguistic, acoustic, and visual cues~\cite{castro-etal-2019-towards,Zadeh_2019_CVPR}.

In this sense, HLSI is built upon the perceptual foundation provided by BLSI.
The development of HLSI should therefore not be viewed as replacing BLSI; rather, it continues to depend on advances in basic semantic perception, which provide reliable clues and grounding for more sophisticated reasoning and generation.

\subsection{Generation Creativity and Logical Coherence}

Creative HLS generation requires models to produce novel content while maintaining semantic consistency, causal plausibility, and alignment with the intended communicative effect. 
Current models, however, remain limited in both creativity and logical coherence, often producing outputs that are either safe and formulaic or novel but semantically disconnected and internally inconsistent~\cite{lei-etal-2025-godbench, song2025generating}. 
This tension is particularly evident in tasks such as humor, metaphor, and narrative generation, where creative effects often arise from unconventional associations but must still remain interpretable within the given context.

Existing reasoning paradigms, such as CoT, primarily organize inference into linear and convergent steps and may therefore be poorly suited to the divergent associations required for creative thinking~\cite{a78fdd6694d54f5491f7393dfc7529e1}. 
Future research should therefore explore new approaches or paradigms that enhance model creativity while preserving the logical correctness and coherence of generated outputs.

%% file: 07_conclusion.tex
\section{Conclusion}

This survey has reviewed the evolution of AI semantic intelligence toward HLSI. 
We have systematically organized this trend from the perspective of semantic complexity, tracing the progression from BLSI, which focuses on explicit, literal, and perceptually grounded meanings, to HLSI, which requires models to understand, generate, and reason about complex, implicit, contextual, and socially situated meanings. 
HLSI aims to enable models to capture higher-level semantics through fine-grained perception and complex reasoning, thereby bringing AI closer to advanced human cognitive capabilities. 
It also has broad application value in human-like interaction, complex reasoning, emotional support, creative generation, AI safety, and other related domains.
This survey provides a comprehensive and in-depth review of existing work on HLS from both understanding and generation perspectives, covering tasks, data, methods, evaluation, and related dimensions. 
The development of HLSI is still in its early stages. 
Significant challenges remain in data scarcity, mechanism-aligned modeling, creative generation, multimodal grounding, and subjective evaluation, etc. 
The path forward requires mining more high-quality HLS data, developing models with fine-grained recognition, complex reasoning, and creative capabilities that can explicitly reason over latent semantic mechanisms, building more reliable benchmarks and evaluation strategies, across real-world contexts.